%% file: iclr2027_conference.tex
\documentclass{article} % For LaTeX2e
\usepackage{iclr2027_conference,times}
\usepackage[T1]{fontenc}
\usepackage{microtype}

\input{math_commands.tex}

\usepackage{float}
\usepackage{hyperref}
\usepackage{url}
\usepackage{xspace}
\newcommand{\methodname}{\mbox{\textsc{PRISM}}\xspace}
\usepackage{graphicx}
\usepackage{wrapfig}
\usepackage{cleveref}
\usepackage{booktabs}
\usepackage{algorithm}
\usepackage{algpseudocode}
\usepackage[most]{tcolorbox}
\usepackage{multirow}
\usepackage{tabularx}

\algtext*{EndIf}

\usepackage{array}
\usepackage{makecell}
\newcolumntype{Y}{>{\centering\arraybackslash}p{2.4cm}}

\usepackage{ragged2e}
\newlength{\eximgwidth}

\newtcbtheorem[
  number within=section
]{promptbox}{Prompt}{
  enhanced,
  breakable,
  colback=gray!5,
  colframe=gray!60,
  fonttitle=\bfseries,
}{prompt}

\title{PRISM: A Category-Theoretic Framework for Measuring and Refining Multimodal Analogies}

\author{Mirella Zeisler, Ojas Shirekar, Mircea Lic\v{a} \& Chirag Raman \\
  Delft University of Technology
}

\iclrfinalcopy % Uncomment for camera-ready version, but NOT for submission.
\begin{document}

\maketitle

\begin{abstract}
Analogical reasoning involves identifying and preserving relational structures across domains. However, existing approaches to AI-driven multimodal analogy generation lack an interpretable measure of whether this structure is understood and maintained in the generated output. We address this gap with Pullback Refinement via Interpretable Structural Mapping (\methodname), a modality-agnostic framework for measuring and improving relational alignment in multimodal analogies, evaluated on visual metaphor generation. \methodname represents analogies as explicit relational mappings grounded in category theory and uses VLMs to instantiate these structures across modalities. Its first component, the pullback score, quantifies relational alignment from the resulting graph representation. On the AnaloBench benchmark, selecting the correct analogy purely by pullback score achieves 82.5\% accuracy, demonstrating that the score captures meaningful relational information. \methodname's second component is an iterative refinement loop that uses
  the pullback score as an in-context feedback signal to iteratively revise the generated image towards
  greater relational depth. VLM-as-a-judge and human evaluations show that \methodname consistently improves
  metaphor consistency and analogy appropriateness over zero-shot generation, with human participants
  preferring the refined output in 57.65\% of pairwise comparisons. However, a qualitative analysis reveals that refinement can favour visually crowded compositions rather than genuinely deeper relational correspondences.
\end{abstract}

\section{Introduction}
\label{sec:introduction}

Human cognition is deeply rooted in analogical reasoning, which is the ability to identify and transfer
relational structures across domains. Already present in young infants, it is central to learning, creativity
and generalisation \citep{gentner_analogy_2017}. An analogy operates not by matching surface-level features,
but instead by aligning the underlying relational structure between a source and a target domain
\citep{gentner_structure-mapping_1983}. This relational abstraction enables humans to apply knowledge across
vastly different contexts, from scientific reasoning to creative expression. As AI systems are increasingly
deployed in open-ended, multi-domain settings, the ability to reason analogically in a similarly
generalisable manner becomes a critical capability.

Advances in large language model (LLMs) and vision-language model (VLMs) capabilities have produced systems
that perform strongly on a wide range of tasks, including several benchmarks designed to probe analogical
reasoning \citep{qin_relevant_2025, musker_llms_2025}. However, strong benchmark performance does not
necessarily imply genuine relational abstraction. There is growing evidence that models exploit statistical
regularities in training data rather than performing true structural inference
\citep{mitchell_abstraction_2021, lewis_evaluating_2024}. This distinction is especially pronounced in the
multimodal setting, where recent work shows that current VLMs still struggle to generalise relational rules
across visual transformations \citep{yiu_kiva_2024}, a difficulty echoed by recent diagnostic benchmarks for
visual analogical mapping \citep{yilmaz_voila_2025}, compositional analogies \citep{du_carv_2026},
and visual metaphor understanding \citep{kundu_imagemet_2025}. Furthermore, even when models produce correct
outputs or generate explanations, their internal reasoning processes remain opaque and do not reliably
correspond to interpretable analogical mappings \citep{chen_can_2025, opielka_analogical_2025}.

\begin{figure}
  \centering
  \includegraphics[width=0.85\linewidth]{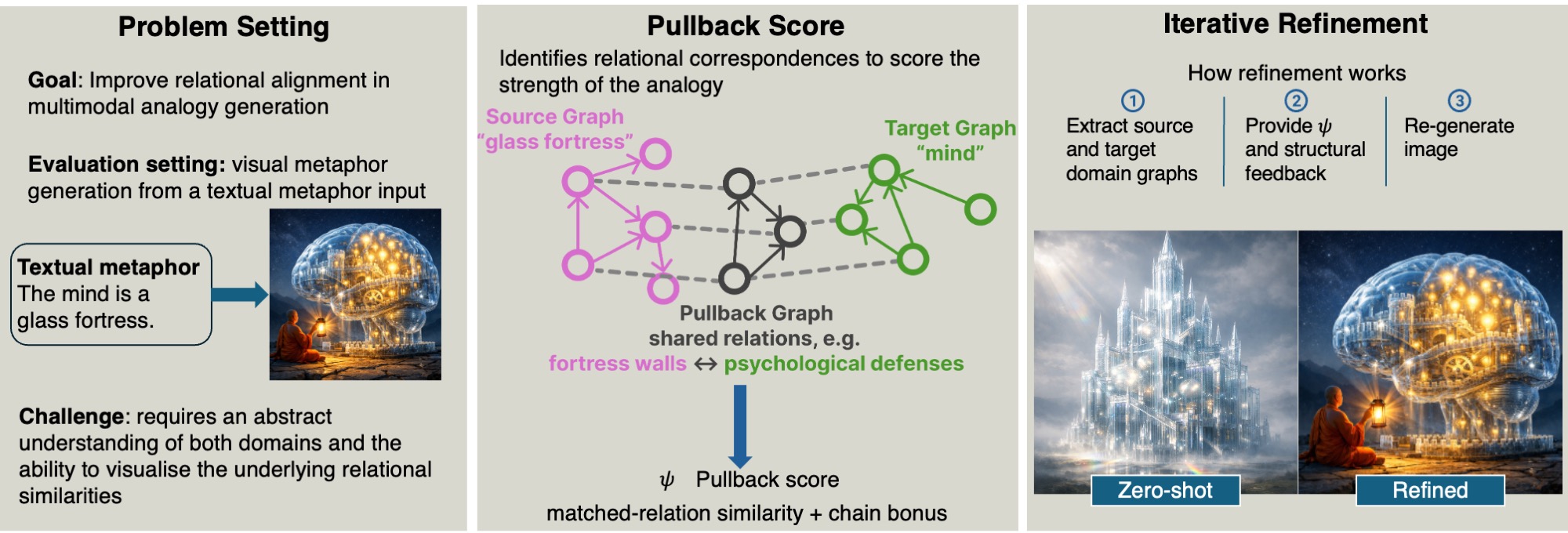}
  \caption{Overview of \textbf{\methodname}, a category-theoretic framework for structural alignment within
    multimodal analogies. The pullback score quantifies relational correspondences between source and target
    domain graphs and provides structural feedback for iterative refinement. We evaluate \methodname on visual
  metaphor generation, where refinement aims to produce analogies with stronger relational correspondences.}
  \label{fig:intro-figure}
\end{figure}

Existing approaches to analogical reasoning in AI tend to be either powerful but opaque, or interpretable but
narrow. Structured prompting strategies and schema-based methods have demonstrated progress in specific task
settings — such as scientific analogy generation \citep{yuan_beneath_2023} or visual metaphor transfer
\citep{xu_beyond_2026} — but these approaches do not generalise across task formulations. What remains
missing is a unified, human-interpretable representation of analogical structure that can operate across
modalities and task types while still leveraging the broad knowledge encoded in state-of-the-art foundation models.

In this paper, we address this gap by proposing Pullback Refinement via Interpretable Structural Mapping
(\methodname), a modality-agnostic structural framework for analogical reasoning that is both
human-interpretable and generalisable across diverse reasoning tasks. \methodname represents analogies as
explicit relational mappings grounded in category-theoretic constraints, and uses VLMs to populate and employ
these representations across multimodal inputs. We evaluate \methodname using visual metaphor generation as
the testbed, since this task requires an abstract understanding of both domains, and generative tasks are
less prone to shortcut learning \citep{mitchell_abstraction_2021, pekar_generating_2020}.

Our contributions are two-fold: (1) We \textbf{translate a category-theoretic definition of analogical
structure into an implementable representational schema}. Building on the formal account of analogies in
\citet{ott_structural_2025}, we instantiate a domain graph representation that realises category-theoretic
analogies within a VLM-based pipeline. Rather than relying on task-specific heuristics, the representation
captures the relational structure between a source and target domain in a unified, modality-agnostic form
while remaining faithful to the underlying theory. From this representation, we derive the \textbf{pullback
score}, a category-theoretic measure of relational alignment that quantifies the extent to which structure is
preserved across a mapping. (2) We \textbf{design and evaluate \methodname's iterative refinement
loop, which improves the relational quality of generated analogies using feedback on structural alignment}.
This loop makes use of the pullback score as a feedback signal to encourage models to generate analogies that
preserve deeper relational correspondences rather than superficial similarities, providing a practical step
towards more transparent, generalisable and structurally grounded analogical reasoning in multimodal AI systems.

\section{Related Work}
\label{sec:related-works}

\textbf{Analogies in AI.} Structure-Mapping Theory characterises analogies as alignments between the relational structures of a base and target domain rather than their surface attributes \citep{gentner_structure-mapping_1983, gentner_analogy_2017}. This idea has been explored through symbolic systems based on hand-crafted representations, neural approaches that learn analogies from data, and neuro-symbolic methods that combine learned representations with explicit structure \citep{hofstadter_copycat_1984, evans_program_1964, marcus_algebraic_2003, mikolov_linguistic_2013, sadeghi_visalogy_2015, mitchell_abstraction_2021, lake_building_2017, evans_learning_2018, yi_neural-symbolic_2018}. More recent work has equipped language and vision-language models with symbolic scaffolds, including tool use and structured reasoning procedures \citep{schick_toolformer_2023, gao_pal_2023, olausson_linc_2023, pan_logic-lm_2023, gupta_visual_2023}. Graph-based approaches make entities, relations, and higher-order correspondences explicit, extending from the Structure-Mapping Engine and visual relational reasoning to learned graph alignment, multimodal knowledge graphs, and structured visual metaphor understanding \citep{falkenhainer_structure-mapping_1989, lovett_modeling_2017, ling_deepgar_2022, zhang_multimodal_2023, xu_beyond_2026}. We extend these directions by introducing a formal graph-theoretic scaffold that VLMs populate and reason over, using soft relational similarity and pullback operations to construct and quantify cross-domain correspondences.

\textbf{Visual Metaphor Generation.} Text-to-image models produce realistic images from prompts \citep{rombach_high-resolution_2022}, but are optimised for literal content rather than the relational structure underlying metaphors, and both VLMs and text-to-image generation models struggle with the abstraction visual metaphors require \citep{akula_metaclue_2023}. Prior work narrows this gap by adding visual detail to prompts \citep{chakrabarty_i_2023}, grounding generation in explicit structures such as attribute-object binding \citep{zhang_gome_2024}, or introducing iterative feedback, either through structured metaphor decomposition and multi-faceted rewards \citep{koushik_minds_2025} or VLM-generated critiques of the visual elaboration \citep{kundu_carve_2026}. \citet{xu_beyond_2026} instead learns from visual references rather than text, transferring metaphors between images. Unlike this prior work, we ground refinement in an explicit, category-theoretic measure of relational alignment rather than a learned or heuristic reward.

\section{Background}\label{sec:bg}
\label{subsec:formalisation}

\begin{wrapfigure}{r}{0.48\textwidth}
  \centering
  \vspace{-0.8\baselineskip}
  \includegraphics[width=\linewidth]{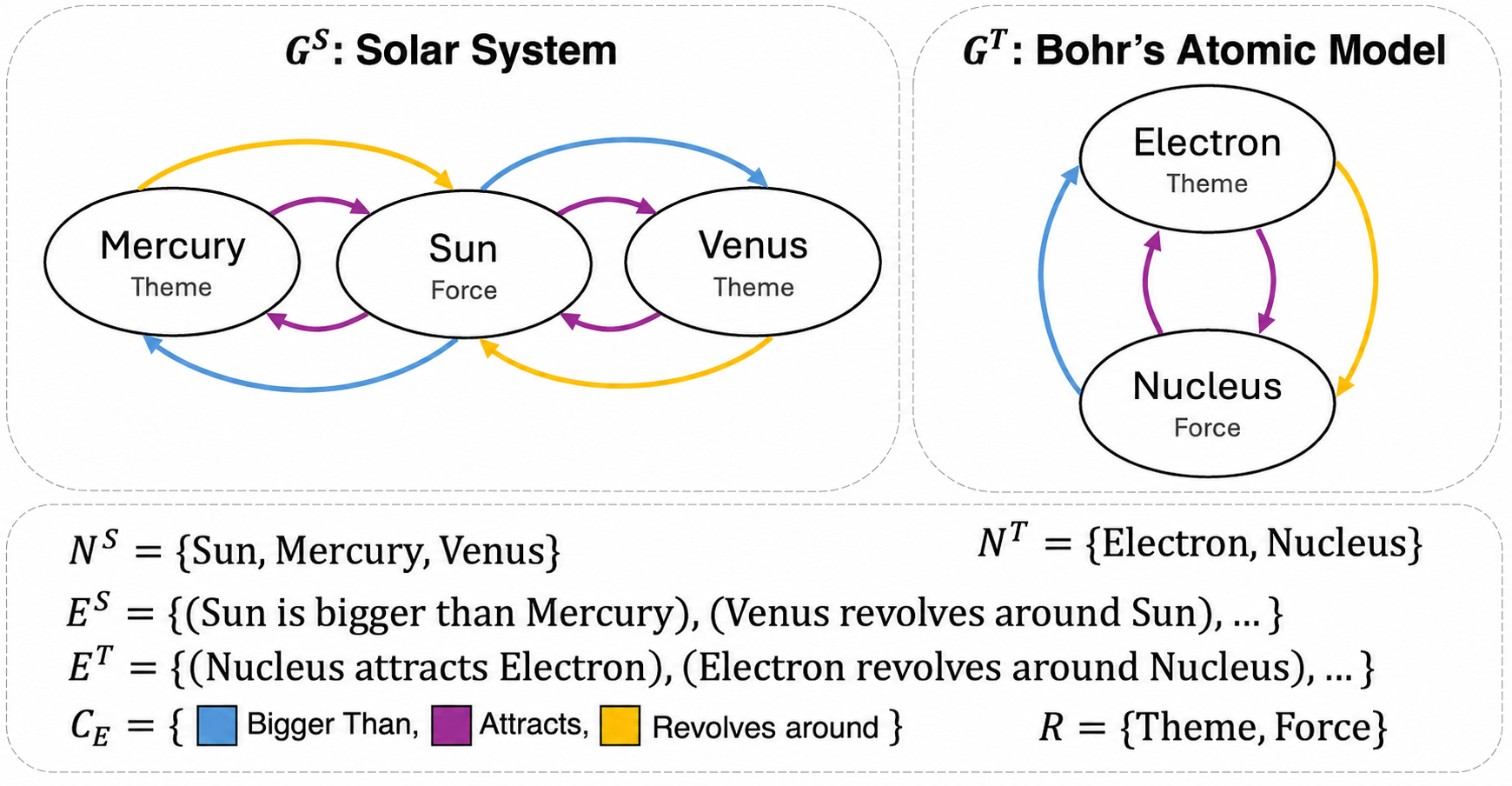}
  \caption{Coloured domain graphs for the solar system (\(G^S\)) and Bohr's atomic model (\(G^T\)). Shared node
  roles and edge colours indicate analogous entities and relations across the two domains.}
  \label{fig:domain-graphs}
  \vspace{-0.6\baselineskip}
\end{wrapfigure}

Category theory studies structure-preserving mappings between collections of objects (see
\citet{awodey_category_2010} and \citet{jencel_category_2021} for a comprehensive introduction).
\citet{ott_structural_2025} applies this to analogies by representing each domain as a coloured multi-graph,
where nodes are the entities of the domain and edges are the relations between them, coloured such that two
edges share a colour exactly when their relations are similar enough to be treated as analogous. The product
of two domain graphs represents every possible pairing between their elements, and the pullback is a
restricted product that additionally requires paired edges to share a colour, so that only
structure-preserving correspondences survive. Individual analogies are then subgraphs of this pullback, and
the preferred mapping is the one that preserves the largest number of relations.

We adopt this representation directly: a domain pair \((D^S, D^T)\) is
represented as a coloured graph pair
\begin{equation}
  G = (G^S, G^T, c_E, c_N), \qquad G^S = (N^S, E^S), \qquad G^T = (N^T, E^T),
  \label{eq:domain-graph}
\end{equation}

where \(N = N^S \cup N^T\) is a set of nodes (entities) and \(E=E^S \cup E^T\) is a set of edges (relations), each
represented as a triple \((u, r, v)\) with \(r\) the relation label. Edges in \(E^S\) are constrained to \(u, v \in
N^S\) and edges in \(E^T\) to \(u, v \in N^T\), so that \(G^S\) and \(G^T\) remain structurally independent prior to
the pullback operation. Every node is assigned a role by \(c_N : N^S \cup N^T \rightarrow R\), drawn from a
semantic-role vocabulary \(R\), and every edge is assigned a colour by \(c_E : E^S \cup E^T \rightarrow C_E\),
drawn from a colour set \(C_E\). 

Formally, sharing \(c_E\) across the two domain graphs gives a cospan of edge-colouring maps into the common
codomain \(C_E\),
\begin{equation}
  E^S \xrightarrow{\ c_E\ } C_E \xleftarrow{\ c_E\ } E^T,
  \label{eq:cospan}
\end{equation}
and the pullback invoked is the pullback of \Cref{eq:cospan} in the category of sets
\begin{equation}
  E^S \times_{C_E} E^T \;=\; \{(e,e') \in E^S \times E^T \mid c_E(e) = c_E(e')\},
  \label{eq:formal-pullback}
\end{equation} together with its projections onto \(E^S\) and \(E^T\). An individual analogy is then a subset \(\mathcal{P} \subseteq E^S \times_{C_E} E^T\) whose endpoints determine a single, consistent element of the node correspondence, computed by the matching procedure in \Cref{alg:greedy}. \Cref{sec:methodology} develops this pullback into a computable score. 

\section{Methodology}
\label{sec:methodology}

We consider a visual metaphor as a source and a target domain \((D^S, D^T)\) related by a shared relational
structure. \Cref{subsec:pullback} constructs a procedure that extracts the coloured graph pair \((G^S, G^T)\)
from a visual metaphor and defines a scalar pullback score \(\psi(G^S, G^T) \in \mathbb{R}_{\geq 0}\) measuring
how much of the target domain's relational structure is preserved in the source domain's depiction.
\Cref{subsec:in-context-refinement} then composes this extraction procedure and score with an
image-generation model and a refinement model into \methodname's iterative refinement loop, which uses \(\psi\)
as feedback to improve subsequently generated visual metaphors.

\subsection{The Pullback Framework for Structural Analogy Extraction and Evaluation}
\label{subsec:pullback}

\begin{figure}[H]
  \centering
  \includegraphics[width=0.8\linewidth]{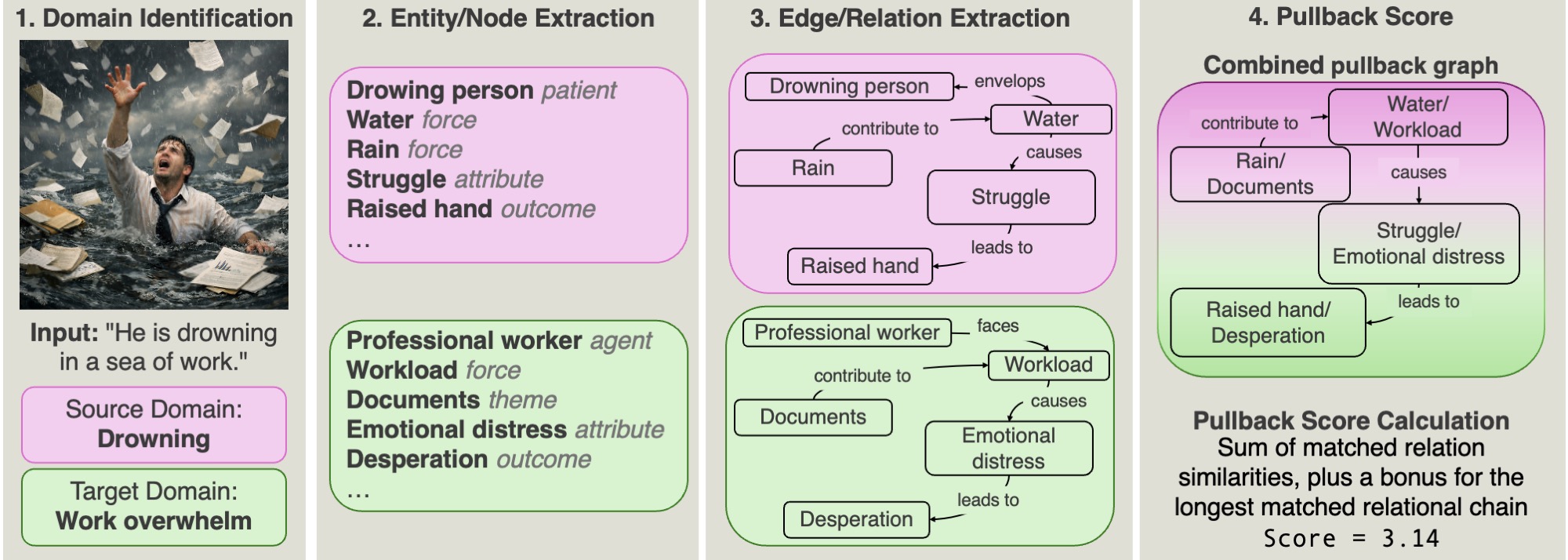}
  \caption{\methodname's four-stage domain-graph extraction pipeline for computing the pullback score from a
    visual metaphor: (1) source and target domain identification, (2) entity extraction, (3) relation
    extraction into coloured domain graphs, and (4) pullback score computation to quantify preserved relational
  structure.}
  \label{fig:extraction-pipeline}
\end{figure}

\paragraph{Constructing coloured domain graphs.}
We formalise the extraction procedure as a map 
\begin{equation}
    \mathrm{ext} : \mathcal{X} \cup \mathcal{M}
    \longrightarrow \mathcal{G} \times \mathcal{G},
    \qquad
    \mathrm{ext}(x) = (G^S, G^T)
    \label{eq:extraction-procedure}
\end{equation}
where \(\mathcal{X}\) is the space of visual metaphors, \(\mathcal{M}\) is the space of textual metaphors, and
\(\mathcal{G}\) is the space of coloured graphs. Given a visual metaphor \(x \in \mathcal{X}\) or a textual
metaphor \(m \in \mathcal{M}\), \(\mathrm{ext}\) returns the coloured graph pair of its two domains. We
instantiate \(\mathrm{ext}\) as a four-stage VLM pipeline that converts a visual metaphor \(x\) into \(G^S\) and
\(G^T\) through incremental abstraction, where each stage produces structured output that conditions the next.
As illustrated in \Cref{fig:extraction-pipeline}, the pipeline identifies the source and target domains \(D^S,
D^T\), extracts their entities into node sets \(N^S, N^T\), and extracts the relations between those entities
into edge sets \(E^S, E^T\) of \((u,r,v)\) triples.

Each extracted entity is also assigned a semantic role (e.g., ``agent'', ``instrument'', or ``outcome''), instantiating the role map \(c_N\). The role inventory is derived from case grammar \citep{fillmore_case_1968} and semantic role labelling resources \citep{kipper_large-scale_2008}. Replacing entity labels with semantic roles reduces the label scatter produced by unconstrained LLM extraction \citep{khojasteh_enhancing_2026}. Relation predicates are likewise drawn from a shared closed vocabulary during extraction (\Cref{appendix:prompts}) for the same reason, to keep the extracted triples well-formed for the sentence encoder. This constrains only how individual edges are labelled, not how they are compared. Cross-domain correspondence does not require these predicate labels to match, since colour-matching is instead computed from the embedding similarity of the full role-substituted triple, so \methodname avoids the fixed, exact-match relation taxonomy used for structural alignment in prior graph-based methods such as SME.

Edge colours are computed by embedding role-substituted relation triples. Specifically, each edge \((u,r,v)\) is rendered as the string ``\(c_N(u)\ r\ c_N(v)\)'', replacing its head and tail entities with their semantic roles (e.g., (\textit{electron}, \textit{revolves around}, \textit{nucleus}) becomes (\textit{theme}, \textit{revolves around}, \textit{force})), ensuring that similarity depends on relational structure rather than lexical overlap. Using \texttt{all-mpnet-base-v2} \citep{reimers_sentence-bert_2019} as the sentence encoder \(\mathcal{F}\), we represent the embedded source and target triples as
\(\hat{E}^S=\mathcal{F}(E^S)\in\mathbb{R}^{|E^S|\times d}\) and
\(\hat{E}^T=\mathcal{F}(E^T)\in\mathbb{R}^{|E^T|\times d}\). We then compute the cosine similarity matrix \(\Phi=\hat{E}^S \cdot(\hat{E}^T)^\top\in\mathbb{R}^{|E^S|\times|E^T|}\), where \(\Phi_{ij}\) denotes the similarity between source edge \(i\) and target edge \(j\). Edges with \(\Phi_{ij}\geq\theta\) are assigned the same colour, instantiating \(c_E\) and replacing SME's exact relation matching with soft semantic equivalence \citep{mikolov_linguistic_2013}. \Cref{alg:greedy} therefore matches edges directly on the thresholded pairwise similarities \(\Phi_{ij}\) rather than on precomputed colour classes, so the pullback score is best understood as a practical relaxation of the categorical construction in \Cref{eq:formal-pullback} rather than an exact instantiation of it. The complete extraction and graph construction procedure is summarised in \Cref{alg:pullback}. We validate the extraction pipeline on the SCAR dataset \citep{yuan_beneath_2023}, with
prompts and evaluation details provided in \Cref{appendix:scar,appendix:prompts}.

\begin{algorithm}
  \caption{Pullback Score Computation}
  \label{alg:pullback}
  \footnotesize
  \begin{algorithmic}[1]
    \Require Source domain graph \(G^S = (N^S, E^S)\), target
    domain graph \(G^T = (N^T, E^T)\), role map
    \(c_N\), sentence encoder \(\mathcal{F}\), matching threshold \(\theta\)
    \Ensure Pullback score \(\psi\), matched pairs \(\mathcal{P}\), coverage \(\rho\)
    \State \(E^S \gets \{(u,r,v)\in E^S \mid u,r,v \text{ all defined}\}\), \(E^T \gets \{(u,r,v)\in E^T \mid
    u,r,v \text{ all defined}\}\)
    \If{\(E^S = \emptyset\) \textbf{or} \(E^T = \emptyset\)}
    \State \Return \(\psi = 0\), \(\mathcal{P} = \emptyset\), \(\rho = 0\)
    \EndIf
    \Statex
    \(\hat{E}^S \gets \mathcal{F}\big(
        \{\text{``}c_N(u)\ r\ c_N(v)\text{''} \mid (u,r,v)\in E^S\}
    \big) \in \mathbb{R}^{|E^S|\times d}\)
    \Comment{Embed role-substituted edges}
    \Statex
    \(\hat{E}^T \gets \mathcal{F}\big(
        \{\text{``}c_N(u)\ r\ c_N(v)\text{''} \mid (u,r,v)\in E^T\}
    \big) \in \mathbb{R}^{|E^T|\times d}\)
    \State \((\psi, \mathcal{P}, \rho) \gets
    \textsc{GreedyPullback}(E^S, E^T, \hat{E}^S, \hat{E}^T, \theta)\)
    \Comment{\Cref{alg:greedy} in \Cref{appendix:algorithms}}
    \State \Return \(\psi\), \(\mathcal{P}\), \(\rho\)
  \end{algorithmic}
\end{algorithm}

\paragraph{The pullback score.}
Given the coloured graphs \(G^S\) and \(G^T\) produced above, the pullback graph contains all pairs of nodes and
edges whose corresponding edges share a colour, thereby filtering the Cartesian product to structurally
consistent correspondences. The resulting score is defined as
\begin{equation}
  \psi(G^S, G^T)=\sigma+0.5\,\ell,
  \qquad
  \sigma=\sum_{(i,j)\in\mathcal P}\Phi_{ij},
  \label{eq:pullback-score}
\end{equation}
where \(\mathcal{P} \subseteq E^S \times E^T\) is the set of source-target edge pairs retained by the pullback,
so pairs \((i,j)\) with colour-matching similarity \(\Phi_{ij} \geq \theta\), accepted greedily in order of
decreasing similarity subject to every accepted pair preserving a single consistent node mapping across the
whole set. Here, \(\ell\) is the length, measured as the number of edges, of the longest cycle-free directed
path in the target graph induced by those matches. The similarity term \(\sigma\) measures relational
correspondence, while the chain bonus rewards metaphors whose correspondences form coherent relational chains
rather than isolated matches. Neither component is arbitrary: we validate both, along with the role
substitution introduced above, on an independent analogy benchmark below.

\paragraph{Validating the pullback score.}
\label{subsec:pullback-validation}
Before using the pullback score as a feedback signal, we validate that it carries meaningful relational
signal on an independent benchmark, separate from the visual metaphor dataset used throughout the rest of
this paper. We use the AnaloBench dataset \citep{ye_analobench_2024}, a collection of stories of varying
length, some of which are analogous. Given a narrative, the task is to identify the analogous story among
four candidates (task T1 by \citet{ye_analobench_2024}).

We compare a zero-shot baseline, in which GPT-4.1 Mini directly selects the analogous story, against extracting domain graphs for the narrative and each candidate and selecting the candidate with the highest pullback score. Zero-shot prompting achieves \(88.5\%\) accuracy. Selecting purely by pullback score achieves \(82.5\%\) accuracy, with a mean reciprocal rank of \(0.895\) and a top-\(2\) accuracy of \(91.5\%\). The correct answer never received a pullback score of zero, indicating that the score always identifies at least some structural correspondence with the true analogy. Thus, although the pullback score is less accurate than zero-shot prompting as a stand-alone decision rule, it captures the structural signal needed for refinement.

The two approaches also make different errors. Of \(200\) examples, both methods agree on \(159\) correct and
\(17\) incorrect cases, while \(6\) are solved only by the pullback score and \(18\) only by zero-shot
prompting (McNemar \(p = 0.0227\)), indicating meaningfully different error distributions. Ablating the score
confirms both of its components are necessary, as removing the chain bonus lowers accuracy to \(76\%\), and
removing the semantic-role abstraction lowers it to \(24\%\), close to the \(25\%\) chance rate on this
four-way task. Extended analysis is provided in \Cref{appendix:analobench}.

\subsection{Feedback-Driven Refinement of Analogies}
\label{subsec:in-context-refinement}

\begin{figure}[H]
  \centering
  \includegraphics[width=0.80\linewidth]{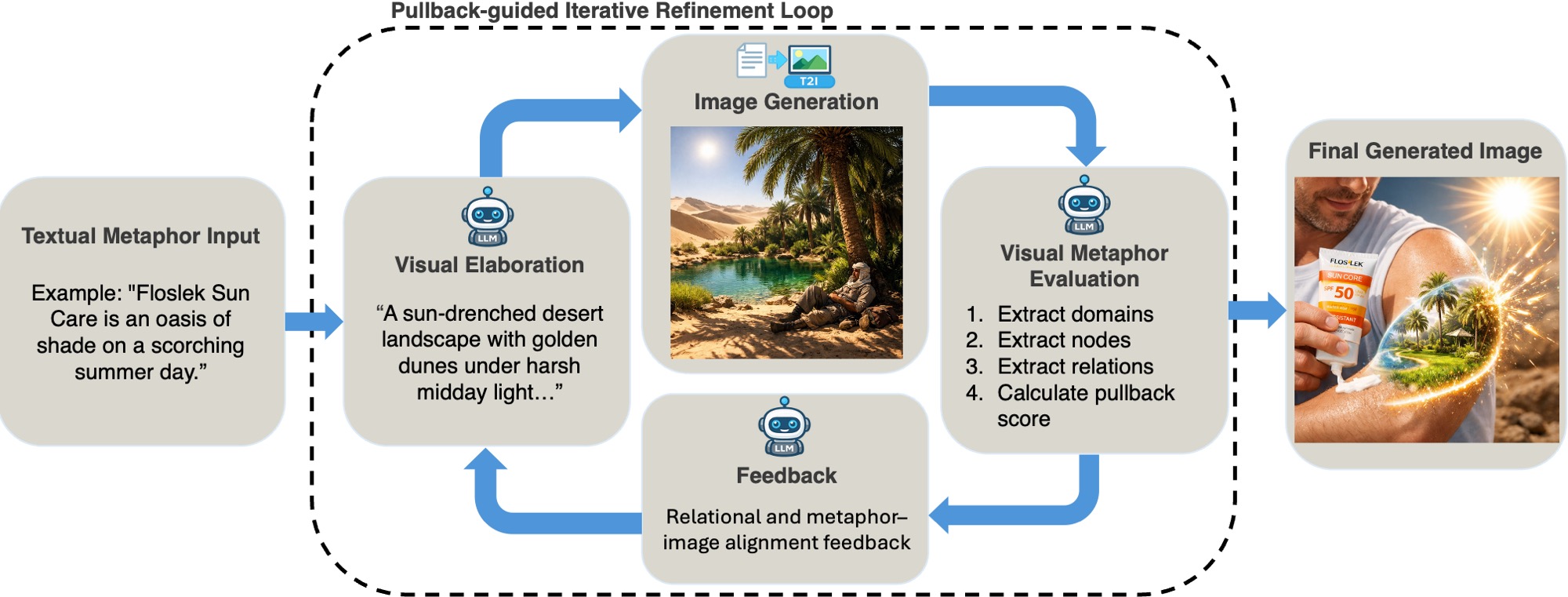}
  \caption{\methodname's refinement loop: a textual metaphor is visualised, scored via the pullback score,
  and revised in-context using feedback from the score and extracted domain graphs, repeating for up to \(k\) rounds.}
  \label{fig:refinement-loop}
\end{figure}

We propose \methodname's refinement loop, which combines an image-generation model \(\mathrm{gen}\), the
extraction procedure \(\mathrm{ext}\), the pullback score \(\psi\), and a refinement model \(\mathrm{ref}\) into
the feedback loop shown in \Cref{fig:refinement-loop}. Given a textual metaphor \(m\) and visual elaboration
prompt \(p\), \(\mathrm{gen}\) produces a visual metaphor \(x\). From \(x\), we extract \((G^S,G^T)=\mathrm{ext}(x)\)
and compute the pullback score \(\psi(G^S,G^T)\). \(\mathrm{ref}\) then revises the prompt using feedback derived
from these signals. Using the pullback score as feedback steers generation towards analogies with stronger
relational correspondences between domains.

Let \(p_0\) denote the initial visual elaboration prompt of the textual metaphor \(m\), produced by a separate
elaboration model. At each refinement round \(t\in \{1,\ldots,k\}\),
\begin{align}
  x_t &= \mathrm{gen}(m,p_{t-1}),
  &&\text{(image generation)}\label{eq:round-gen}\\
  (G^S_t,G^T_t) &= \mathrm{ext}(x_t),\qquad
  \psi_t=\psi(G^S_t,G^T_t),
  &&\text{(structural evaluation)}\label{eq:round-eval}\\
  p_t &= \mathrm{ref}(m,p_{t-1},f_t),
  &&\text{(prompt refinement),}\label{eq:round-refine}
\end{align}
where the refinement feedback at round \(t\) is \(f_t=\{\psi_t,\rho_t,\mathcal{P}_t,G^S_t,G^T_t,a_t\}\). Here,
\(\psi_t\) is the pullback score, \(\rho_t=|\mathcal{P}_t|/|E_{T,t}|\) is the edge coverage, \(\mathcal{P}_t\) is
the set of matched edge pairs, \(G^S_t\) and \(G^T_t\) are the extracted source and target graphs, and \(a_t\) is
the CLIPScore \citep{hessel_clipscore_2021} image--text alignment between \(x_t\) and \(m\).

% \begin{figure}[H]
%     \centering
%     \includegraphics[width=0.88\linewidth]{refinement-loop.jpg}
%     \caption{\methodname's refinement loop: a textual metaphor is visualised, scored via the pullback
% score, and revised in-context using feedback from the score and extracted domain graphs, repeating for up
% to $k$ rounds. Notation: $t$, current refinement round; $k$, maximum number of rounds; $\psi_t$, pullback
% score of round $t$'s image; $\varepsilon$, convergence threshold for early stopping.}
%     \label{fig:refinement-loop}
% \end{figure}

Together with the original metaphor \(m\) and previous elaboration \(p_{t-1}\), these signals provide both global
and local feedback. The scalar scores \(\psi_t\), \(\rho_t\), and \(a_t\) summarise structural quality, relational
coverage, and semantic alignment, while \(\mathcal{P}_t\) and the extracted graphs identify which target-domain
relations remain unmatched. The refinement model is instructed to preserve correctly expressed structure and
revise only the missing or inconsistent relations. The process terminates when either \(t=k\) or the pullback
score converges, i.e. \(|\psi_t - \psi_{t-1}| < \varepsilon\), at which point the final image \(x_t\) is returned.

\section{Experimental Setup}
\label{sec:experimental-setup}

\textbf{Baselines.}
We evaluate \methodname's generated analogies against two classes of baselines. First, we compare the refined
images against the zero-shot outputs produced by Flux.2 Klein 4B
\citep{black_forest_labs_black-forest-labsflux2-klein-4b_2026}, Stable Image Core v1.1
\citep{stability_ai_stability_2024}, GPT Image 1.5 \citep{openai_gpt_2025} and Ideogram 4 \citep{ideogram2025}. Second, we compare
\methodname's refinement procedure against GEPA \citep{agrawal_gepa_2025}, a reflective prompt optimiser that
iteratively proposes, evaluates, and selects prompt variants using natural-language feedback and Pareto-based
evolutionary search. We evaluate GEPA with two different optimisation objectives, one that uses the pullback
score, while the other uses CLIPScore \citep{hessel_clipscore_2021}.

\textbf{Dataset.}
The dataset contains two subsets, the first consists of \(125\) human-created visual metaphors
collected from online sources and existing datasets \citep{hussain_automatic_2017, hwang_memecap_2023},
including advertisements (\(83\)), cartoons and artworks (\(18\)), awareness campaigns (\(20\)), and memes
(\(4\)). For each image, we manually transcribe the underlying metaphorical intent into a textual metaphor
and annotate the corresponding source and target domains. The second subset contains \(125\) textual
metaphors from the HAIVMet dataset \citep{chakrabarty_i_2023}. We publicly release the collected and
annotated human-created metaphors, together with the visual metaphors generated for the full set of textual
metaphors used in our experiments, at \url{https://zenodo.org/records/21702355}.

\textbf{Metrics.} \label{subsec:metrics}
Evaluating visual metaphors requires both understanding the intended metaphorical meaning and recognising the abstract source and target domains depicted in the image. We therefore rely on both automated and human judgements. We evaluate outputs using two VLM judges, Claude Opus 4.8 \citep{anthropic_introducing_2026} and GPT-5.2 \citep{openai_introducing_2025}, and corroborate these scores with human ratings. Including GPT-5.2 provides an independent model family because the refinement loop uses Claude Sonnet 4.6. Each model is prompted to score each visual metaphor on three 10-point scales: \emph{metaphor consistency (MC)}, measuring whether the image preserves the core logic of the textual metaphor, \emph{analogy appropriateness (AA)}, assessing the validity of the functional and formal correspondences between the depicted domains, and \emph{conceptual integration (CI)}, evaluating whether the two domains are fused naturally and coherently \citep{xu_beyond_2026}. The evaluation prompts adapt \citet{xu_beyond_2026} to assess generated images against textual metaphors rather than source images. We further conduct a human evaluation with 80 participants, in which each metaphor is independently reviewed by two participants. Each participant scored five pairs of zero-shot and refined metaphors on analogy appropriateness and metaphor consistency, and selects which image better captures the underlying relational structure. The full survey is provided in \Cref{appendix:user-study}.

\textbf{Implementation.}
After comparing three state-of-the-art image generator models in a zero-shot setting (see
\Cref{appendix:zero-shot-examples,appendix:zero-shot-quant}), we selected GPT Image 1.5
\citep{openai_gpt_2025} as the primary image generator. We conduct hyperparameter tuning to select the base
VLM used for visual elaboration and pullback extraction, to define the edge similarity matching threshold
\(\theta\), and to calibrate the refinement-loop parameters \(k\) and \(\varepsilon\)
(\Cref{appendix:hyperparameters}). This procedure selects Claude Sonnet 4.6
\citep{anthropic_introducing_2026} as the feedback model, with \(k = 10\), \(\varepsilon = 0.05\), and a matching
threshold of \(\theta = 0.40\). To test the applicability of the methodology with different models, we later re-ran the experiments with GPT-6 Astra \citep{openai2026gpt6astra}.

\begin{wrapfigure}[24]{r}{0.30\textwidth}
  \centering
  \includegraphics[width=\linewidth]{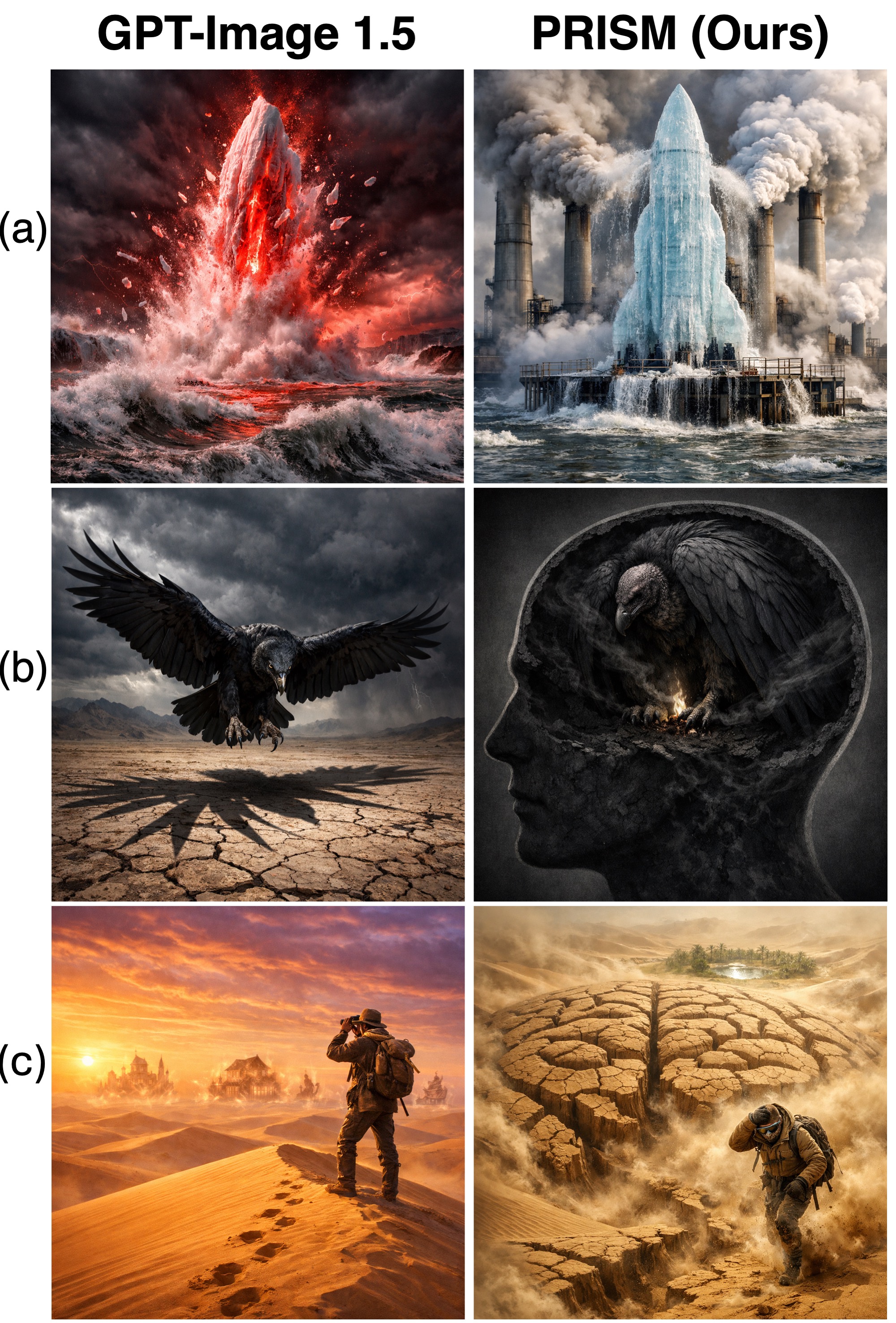}
  \caption{Qualitative comparison of generated visual metaphors. (a) "Global ice melting is as dangerous as a missile." (b) "Thought is a vulture" (c) "My mind is a desert and I am an explorer"}
  \label{fig:comp-with-baselines}
\end{wrapfigure}

\section{Results}
\label{sec:results}

\paragraph{Qualitative comparisons show that \methodname improves the recognisability of both domains and strengthens the relational correspondences between them.} As shown in \Cref{fig:comp-with-baselines}, state-of-the-art image-generation models often fail to depict both domains of a metaphor clearly in zero-shot generation, particularly when one domain involves an abstract concept such as ``mind'' or ``thought''. \methodname produces more interpretable visual metaphors by making both domains recognisable and rendering their relationship more explicit. In example (a) from \autoref{fig:comp-with-baselines}, for instance, the refined image introduces smokestacks alongside the melting ice, visually linking industrial emissions to global ice loss and thereby clarifying the intended causal correspondence. This qualitative improvement is also reflected in the human evaluations, in which participants preferred the \methodname-refined image in \(57.65\%\) of forced-choice comparisons. Examples across all zero-shot generations and PRISM with different refinement models can be found in \autoref{appendix:qual-comparisons}.

\paragraph{\methodname significantly improves metaphor consistency and analogy appropriateness compared to zero-shot image generation.} As \Cref{tab:refinement-comparison} shows, \methodname outperforms zero-shot visual metaphor generation across all axes in both human and VLM evaluations. Notably, analogy appropriateness and metaphor consistency achieve significantly larger gains than conceptual integration. This indicates that refinement acts as a semantic corrector rather than an aesthetic one, strengthening the metaphor's conceptual
relations without necessarily improving overall image quality or composition. We also tested GPT-6 Astra as an alternative extraction and refinement model. Metaphor consistency, analogy appropriateness, and conceptual integration all improve on the Astra runs, but by smaller margins than with Claude Sonnet 4.6. This may be because the Astra zero-shot scores are already higher, so there is less room for improvement. 

\begin{wrapfigure}{r}{0.49\columnwidth}
  \centering
  \vspace{-0.8\baselineskip}
  \includegraphics[
    width=\linewidth,
    trim={0 0 0 0},
    clip
  ]{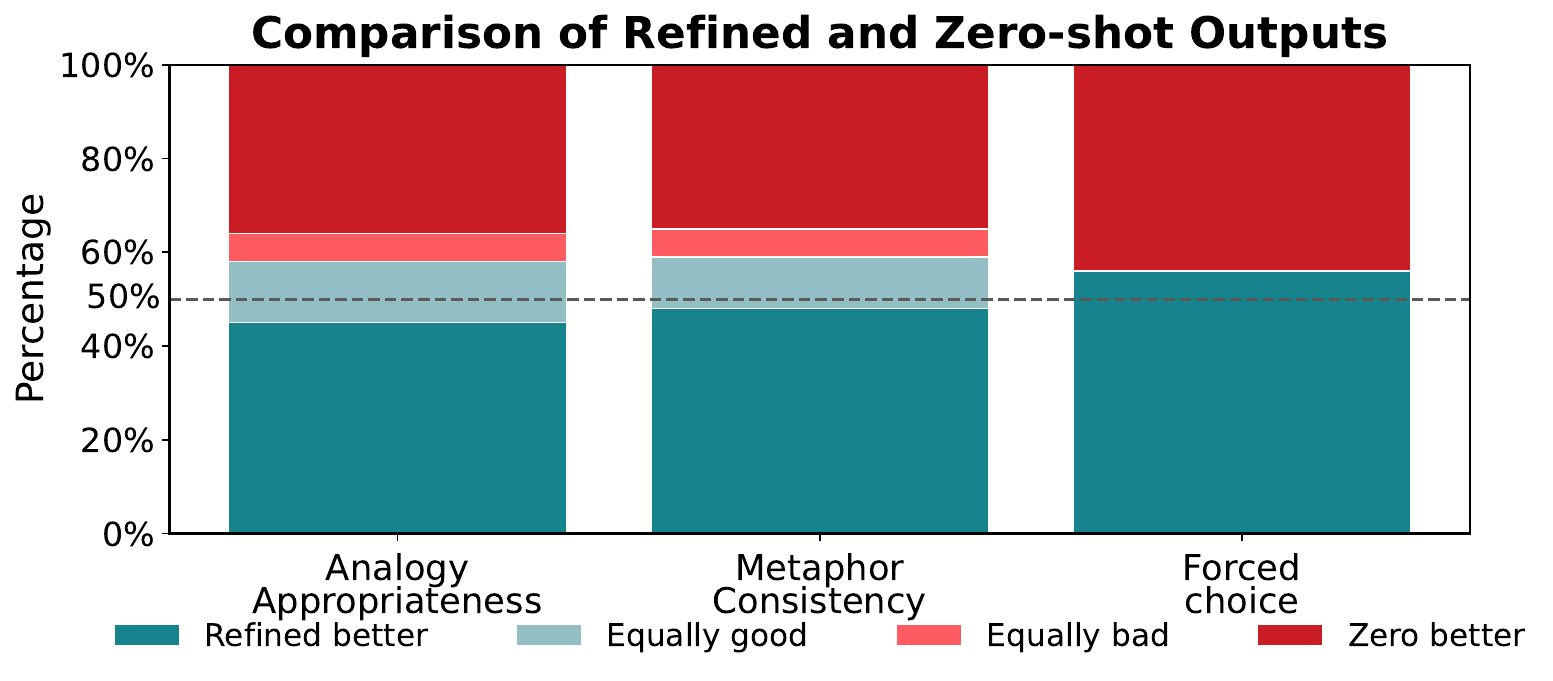}
  \caption{Human pairwise comparison of refined and zero-shot outputs. Tied ratings are classified as equally good for scores of at least 5 and equally bad otherwise.}
  \label{fig:human-comparison}
  \vspace{-0.8\baselineskip}
\end{wrapfigure}

\Cref{fig:human-comparison} depicts how the scalar ratings between the zero-shot and refined images compare across the evaluation axes and shows that refined images are rated higher more often than zero-shot images. Across both axes, the combined proportion of ‘refined better’ and ‘equally good’ outcomes exceeds \(50\%\). Ties in quality between zero-shot and refined are mostly favourable, but there is also a substantial proportion of metaphors in which zero-shot generation outperforms the refined metaphors.

\begin{table}[H]
  \centering

  % Refinement comparison
  \begin{minipage}[t]{0.38\textwidth}
    \centering

    \caption{Zero-shot (ZS) versus \methodname-refined (Ref.) outputs, scored 0 to 10. VLM scores average all judges; human raters did not assess CI.
      \(\Delta\) is the mean improvement.}
    \label{tab:refinement-comparison}

    \vspace{\baselineskip}
    \footnotesize
    \setlength{\tabcolsep}{3pt}

    \begin{tabular}{@{}llccc@{}}
      \toprule
      \multicolumn{2}{@{}l}{\textbf{Model} / \textit{Judge}}
        & \textbf{MC} & \textbf{AA} & \textbf{CI} \\
      \midrule
      Sonnet 4.6     & ZS         & 6.42 & 6.36 & 7.85 \\
      \textit{VLM}   & Ref.       & 7.65 & 7.44 & 8.15 \\
                     & \(\Delta\) & \(\uparrow\)1.24 & \(\uparrow\)1.08 & \(\uparrow\)0.30 \\
      \midrule
      Sonnet 4.6     & ZS         & 5.48 & 5.27 & -- \\
      \textit{Human} & Ref.       & 5.93 & 5.84 & -- \\
                     & \(\Delta\) & \(\uparrow\)0.45 & \(\uparrow\)0.57 & -- \\
      \midrule
      GPT-6 Astra    & ZS         & 7.65 & 7.33 & 7.27 \\
      \textit{VLM}   & Ref.       & 8.18 & 7.81 & 7.32 \\
                     & \(\Delta\) & \(\uparrow\)0.53 & \(\uparrow\)0.49 & \(\uparrow\)0.05 \\
      \bottomrule
    \end{tabular}
  \end{minipage}
  \hfill
  %
  % Baselines and ablations
  \begin{minipage}[t]{0.595\textwidth}
    \centering

    \caption{Comparisons between \methodname, zero-shot (ZS) image-generation
      baselines, GEPA \citep{agrawal_gepa_2025}, and ablations of \methodname.
      ``Overall'' averages scores across all axes and both judges.
      Best result per column is shown in bold.}
    \label{tab:baseline-comparisons}

    \vspace{\baselineskip}
    \footnotesize
    \setlength{\tabcolsep}{2.5pt}

    \begin{tabular}{@{}lccc|ccc|c@{}}
      \toprule
      & \multicolumn{3}{c|}{\textbf{Opus 4.8}}
      & \multicolumn{3}{c|}{\textbf{GPT-5.2}}
      & \multirow{2}{*}{\textbf{Overall}} \\
      \cmidrule(r){2-4}
      \cmidrule(lr){5-7}
      \textbf{Condition}
      & \textbf{MC} & \textbf{AA} & \textbf{CI}
      & \textbf{MC} & \textbf{AA} & \textbf{CI}
      & \\
      \midrule
      Stable Core (ZS)
      & 3.75 & 3.70 & 4.90
      & 4.30 & 4.95 & 6.40
      & 4.67 \\
      Flux Klein (ZS)
      & 5.05 & 4.90 & 6.06
      & 5.16 & 5.63 & 7.32
      & 5.69 \\
      GPT-Image (ZS)
      & 6.25 & 6.15 & 7.50
      & 6.50 & 6.70 & 7.85
      & 6.83 \\
      Ideogram 4 (ZS)
      & 5.26 & 5.37 & 6.84
      & 5.37 & 6.05 & 7.24
      & 6.02 \\
      \midrule
      GEPA (pullback)
      & 6.57 & 6.29 & 6.34
      & 7.02 & 6.62 & 6.88
      & 6.62 \\
      GEPA (CLIPScore)
      & 6.78 & 6.35 & 6.46
      & 7.23 & 6.78 & 6.93
      & 6.76 \\
      \midrule
      No domain guidance
      & 4.32 & 4.45 & 4.92
      & 4.88 & 5.15 & 5.87
      & 4.93 \\
      One-shot extraction
      & 6.24 & 6.23 & 6.03
      & 6.75 & 6.78 & 6.80
      & 6.47 \\
      \midrule
      \textbf{\methodname (Ours)}
      & \textbf{7.60} & \textbf{7.30} & \textbf{8.35}
      & \textbf{7.85} & \textbf{7.70} & \textbf{8.43}
      & \textbf{7.87} \\
      \bottomrule
    \end{tabular}
  \end{minipage}
\end{table}

\paragraph{Evaluating visual metaphors is inherently difficult and subjective.} Agreement is limited both between human and automated evaluators and among human raters themselves. Human and VLM scores are significantly positively correlated, but only moderately for zero-shot images (Pearson's \(r=0.439\)--\(0.446\)) and more weakly for refined images (\(r=0.270\)--\(0.303\)). Similarly, agreement between human and VLM forced-choice preferences is only fair (Cohen's \(\kappa=0.270\), 66.2\% raw agreement), with the VLM identifying 81.4\% of human preferences for refined images but only 44.4\% of preferences for zero-shot images. Human inter-rater agreement is also low for absolute scores, with Krippendorff's \(\alpha=0.082\) for analogy appropriateness and \(\alpha=0.111\) for metaphor consistency, but increases for forced-choice judgements (\(\alpha=0.233\)). Nevertheless, raters agree on the direction of improvement for 71.5\% of metaphors in analogy appropriateness and 74.7\% in metaphor consistency, suggesting that relative comparisons are more reliable than absolute quality ratings. Together, these results show that visual metaphor evaluation depends on subjective interpretations of whether relational structure has been preserved, motivating the pullback score as a directly computable and less rater-dependent measure of relational alignment.

\paragraph{Explicit domain guidance and structured refinement drive \methodname's performance.} We conduct baseline and ablation comparisons. As shown in \Cref{tab:baseline-comparisons}, \methodname achieves the highest overall score of 7.87, outperforming all zero-shot baselines, GEPA variants, and ablations. Among the GEPA baselines, optimisation with CLIPScore performs best, reaching an overall score of 6.76, compared with 6.62 when GEPA is optimised directly with the pullback score. Both remain below the GPT-Image zero-shot baseline at 6.83 and substantially below \methodname. Replacing \methodname's multi-stage extraction with a single extraction step reduces performance to 6.47, while removing domain guidance from the refinement feedback, including both the input metaphor and CLIPScore, produces the largest degradation, to 4.93 overall. These ablations show that \methodname benefits from explicit source–target grounding and iterative structured refinement. GEPA performs better with CLIPScore than with the pullback score, suggesting that image–text alignment is a more effective prompt-level objective in this setting. However, neither variant matches the full pipeline, and GEPA with CLIPScore lacks \methodname’s modality-agnostic structural representation.

\begin{figure}[H]
  \centering
  \begin{minipage}[t]{0.35\linewidth}
    \centering
    \includegraphics[width=\linewidth]{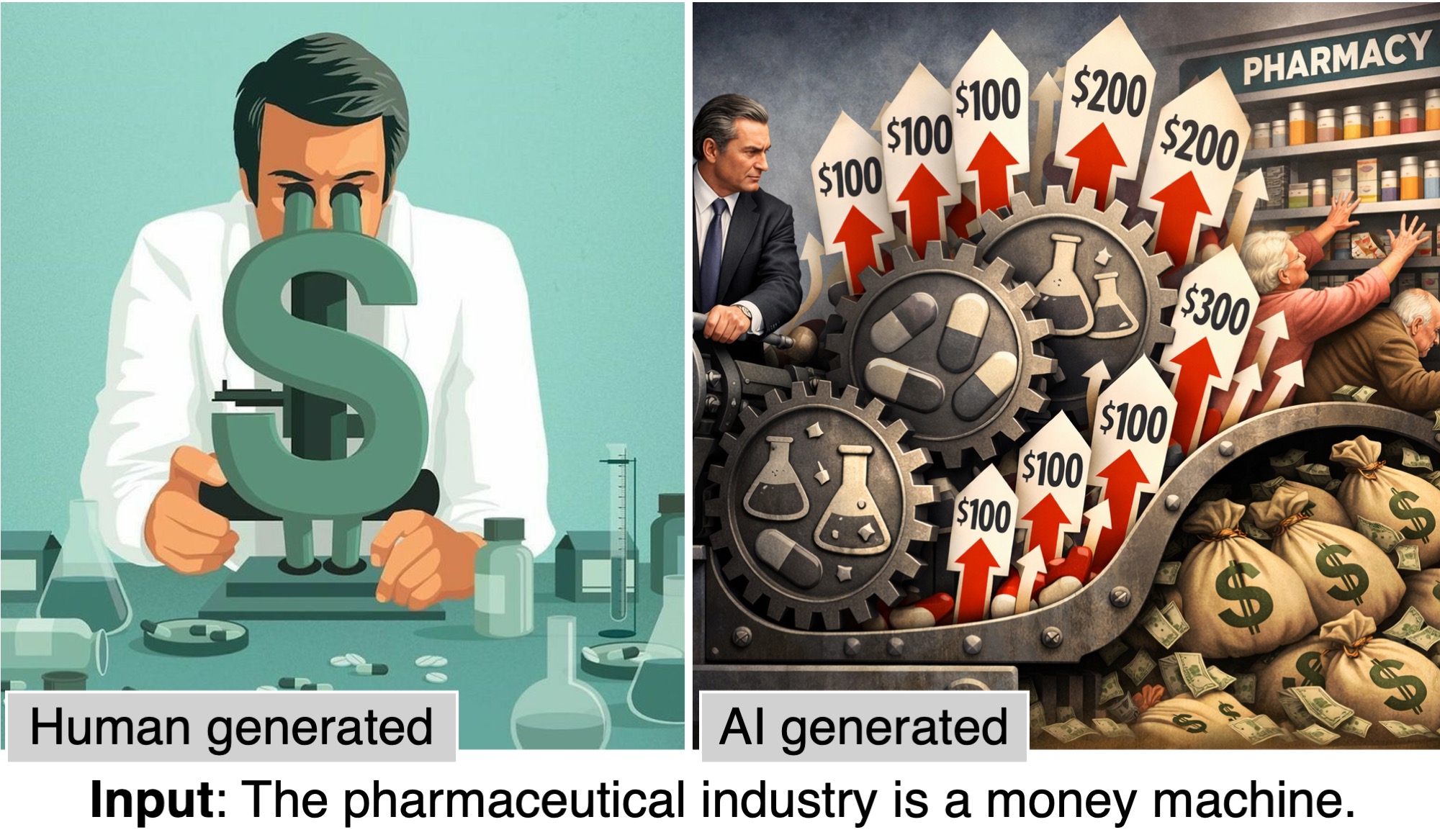}
  \end{minipage}
  \begin{minipage}[t]{0.37\linewidth}
    \centering
    \includegraphics[width=\linewidth]{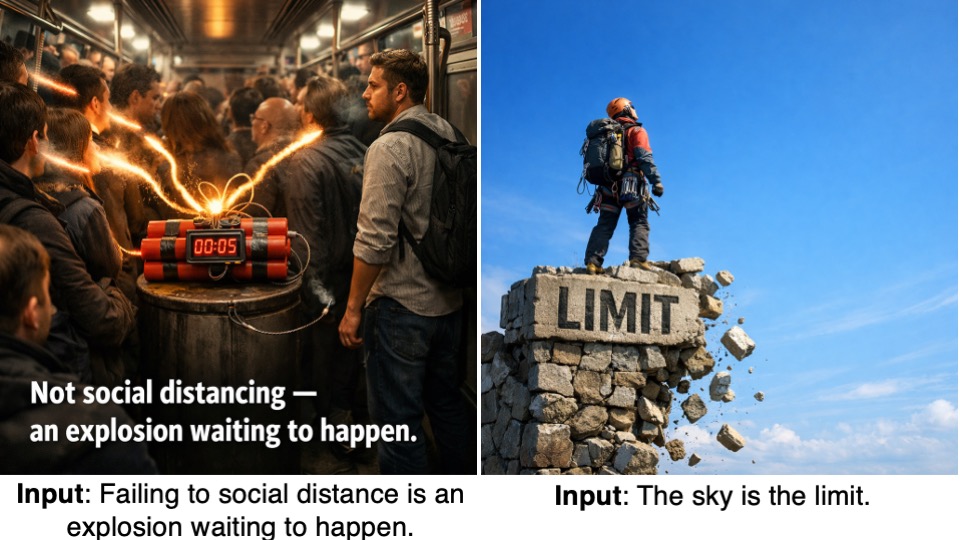}
  \end{minipage}
  \caption{Representative failure modes. \textbf{(a) Maximalism:} whereas the human-designed metaphor \citep{schmitz_nih_2025} communicates pharmaceutical corruption through a single dollar-shaped microscope, the refined output introduces multiple redundant monetary cues. \textbf{(b) Text as a shortcut:} generated images use written labels to identify domains such as ``social distancing'' or ``limit'', improving domain
  extraction without strengthening the underlying relational correspondence.}
  \label{fig:limitation-examples}
\end{figure}

\paragraph{Limitations: maximalism and text as a shortcut.} Qualitative inspection suggests two recurring failure modes. First, refinement often produces increasingly cluttered compositions by adding rather than removing visual elements. Second, generated images may rely on literal text to establish the intended domains; because the VLM-based extractor can read this text directly, such images may receive a high pullback score without expressing a clearer visual correspondence, as illustrated in \Cref{fig:limitation-examples}.

\section{Conclusion}
\label{sec:conclusion}

This paper introduced \methodname, comprising the pullback score as a category-theoretic measure of relational alignment computed from VLM-extracted domain graphs, and a feedback-driven refinement loop that uses it to guide iterative, test-time image generation. We validated that the pullback score carries meaningful analogical signal on the AnaloBench benchmark, and showed that VLMs struggle to depict the domains and relational correspondences of a textual metaphor zero-shot, a dissociation \methodname is designed to close. Across 200 metaphors, refinement yields statistically significant improvements in metaphor consistency and analogy appropriateness according to VLM judges and human raters. The direction of the improvement is replicated when running the loop using GPT-6 Astra, but smaller gains highlight the dependence of the pipeline’s performance on the models used for image generation, extraction, and refinement. At the same time, our evaluations show that judging analogical quality, whether by VLM or by human, is itself difficult and only moderately consistent, highlighting the need for interpretable, structurally grounded metrics like the pullback score.

Beyond visual metaphor generation, we see this work as a step towards making analogical reasoning in multimodal AI systems more transparent, measurable, and improvable, rather than treating relational alignment as an unverifiable byproduct of generation. Applying \methodname beyond visual metaphor generation in future research, to other multimodal or purely textual analogical reasoning tasks, would test how far this
structural approach to relational alignment generalises.

\subsection*{AI use statement}

In this work, we used generative AI tools for helping develop theoretical models or conceptual frameworks,
formulating mathematical claims, designing or providing feedback on research  methodology or experiments,
implementing methods and interpreting results. We have not used generative AI tools for proposing or refining
hypotheses, generating synthetic data sets, cleaning and reformatting datasets, or supporting qualitative and
thematic data analysis. Providing critical ingredients for proving mathematical claims, assisting in the
writing of proofs, and assisting with translation are not applicable to this work.
Additionally, we used generative AI tools for modifying scientific figures, creating or editing software
code, editing the research paper to improve readability, summarizing existing literature and proposing a
title for the research paper. We have reviewed all AI-assisted work.

The conceptual framework was grounded in the category-theoretic understanding of analogies in
\citet{ott_structural_2025} and generative AI assisted in translating that formal account into an
implementable graph schema. The pullback score in
\Cref{eq:pullback-score} and the matching procedure in \Cref{alg:greedy} were formulated with AI assistance
and verified by the authors with hand-crafted examples adapted from the SCAR dataset
\citep{yuan_beneath_2023}. AI-generated code was reviewed by the authors and tested against expected
behaviour, e.g. the extraction pipeline was independently verified on the SCAR dataset
\citep{yuan_beneath_2023} and the AnaloBench dataset \citep{ye_analobench_2024}.

The figures in the paper are hand-crafted by the authors, and only modified using generative AI. Two such
examples are \Cref{fig:domain-graphs}, where the curved arrows were added using AI tools, and
\Cref{fig:refinement-loop}, which contains AI-generated icons for LLMs and T2I generation models. For the
literature review, AI was used to collect and summarize relevant literature, but each cited work was
consulted by at least one author to confirm that they exist and to validate the claims attributed to them.
All report sections were initially drafted by the authors, and refined for readability and clarity using
generative AI tools. We take responsibility for the final content
of this work, including text, claims or artifacts produced with the aid of
generative AI.

\subsection*{Ethics statement}

\paragraph{Human subjects.} The human evaluations reported in \Cref{sec:experimental-setup} was reviewed and
approved by our institute's Human Research Ethics Committee before approaching participants, and reviewed by
the faculty's data management steward. Participants were recruited through a third-party survey exchange platform and compensated in platform credit. Furthermore, participation was voluntary with the option to
withdraw at any time. Beyond the visual metaphor evaluations, we collected only age, gender and education
levels, and participants also had the option to skip these questions. All images shown to participants were
screened beforehand and visual metaphors that were deemed violent or disturbing were removed. However, we
also acknowledge the limitations of the resulting sample. Recruitment through a survey exchange platform
leads to a skewed pool of participants that is largely focused on students and researchers. Because metaphor
interpretation is subjective and culturally situated, this sample should not be taken as a representation of
the general audience.

\paragraph{Data provenance and licensing.} In our evaluation, we draw our dataset from two main sources, as
described in \Cref{sec:experimental-setup}. 125 textual metaphors are collected from the HAIVMet dataset by
\citet{chakrabarty_i_2023} and the 125 human-created visual metaphors used to inform the remaining textual
metaphors were collected from online sources and existing research datasets \citep{hussain_automatic_2017,
hwang_memecap_2023}. As to not violate these third-party copyrighted works, these collected visual metaphors
are not redistributed in the online dataset found at \url{https://zenodo.org/records/21702355}. Instead,
this dataset contains only the transcribed textual metaphors, the annotated source and target domains, and
the source URLs of the images. Researchers wishing to reproduce our results on the human-generated subset can
obtain the source images from their original providers under those providers' terms.

\subsection*{Reproducibility statement}

We have aimed to make \methodname reproducible from the paper by including the algorithm for the pullback
score computation in \Cref{alg:pullback} and \Cref{alg:greedy} and the prompts for the domain graph
extraction pipeline in \Cref{appendix:prompts}. \Cref{sec:methodology} furthermore specifies the sentence
encoder used for the edge colouring and the used hyperparameter values are defined in
\Cref{sec:experimental-setup} and \Cref{appendix:hyperparameters}. 

However, there are also a few factors that limit the reproducibility of the results. First of all, every
stage of our pipeline is probabilistic, because image generation offers no seed control and domain-graph
extraction varies across repeated calls on the same image. Additionally, most of our experiments use
API-based proprietary models, which might be deprecated in favour of newer versions.



\bibliography{iclr2027_conference}
\bibliographystyle{iclr2027_conference}

\newpage
\appendix

\section{Prompts}
\label{appendix:prompts}

\subsection{Extraction Pipeline Prompts}

\begin{promptbox}
  {Stage 1: Domain Identification}
  {prompt:domain-identification}
  STAGE 1 of dual-domain graph extraction from a visual metaphor.

  YOU ARE GIVEN: a single image that depicts a visual metaphor. A visual metaphor combines two distinct
  conceptual domains:
  \begin{itemize}
    \item the TARGET domain (the abstract concept the metaphor is about) and
    \item the SOURCE domain (the concrete concept the target is being compared to).
  \end{itemize}

  For example, an image of a single man wreathed in fog is a visual metaphor where the TARGET is
  ``loneliness" (abstract, what the image is about) and the SOURCE is ``fog" (concrete, what loneliness is
  being compared to).

  YOUR TASK: identify the two domains depicted.

  YOUR OUTPUT WILL BE USED BY: a downstream pipeline that extracts a graph of nodes (entities) and edges
  (relations) for each domain independently, then computes a categorical pullback between the two graphs to
  score the structural strength of the metaphor. So the domain names you return must be:
  \begin{itemize}
    \item single canonical noun phrases, lowercase, no articles
    \item abstract enough to admit multiple instantiations (prefer ``loneliness" over ``the lonely old man's
      feelings"; prefer ``fog" over ``the rolling grey morning fog"); however, if the domain clearly refers
      to a specific entity (e.g. ``Eiffel Tower" or ``Mercedes Benz"), return that as the domain name.
    \item No longer than 2-3 words (e.g. ``social isolation" is fine, but ``the feeling of being socially
      isolated in a crowd" is not)
    \item distinct from each other (target != source)
  \end{itemize}

  If the image does not clearly depict a visual metaphor (e.g. it is a literal photograph with no second
  domain implied), still return your best guess for what the dominant depicted concept is as the target, and
  ``none" for the source. Be critical here: if you cannot clearly identify the two domains that the metaphor
  is trying to depict, return ``none" for the source domain. The downstream pipeline will detect this.
\end{promptbox}

\newpage

\begin{promptbox}
  {Stage 2: Node Extraction}
  {prompt:node-extraction}
  STAGE 2 of dual-domain graph extraction from a visual metaphor.

  YOU ARE GIVEN:
  \begin{itemize}
    \item the same image as Stage 1
    \item the target domain and source domain identified in Stage 1
  \end{itemize}

  YOUR TASK: list the key entities (nodes) for each domain. Each node has two parts:
  \begin{enumerate}
    \item A `name`: a concrete, image-grounded label for what this entity actually is in the depicted scene
      or in the abstract domain. Names should be specific to the domain and faithful to what is visible or
      genuinely implied by the image. Examples:
      \begin{itemize}
        \item source domain ``construction site": construction\_worker, hammer, nail, concrete\_block, steel\_beam
        \item target domain ``creative work": writer, idea, manuscript, struggle, vision
      \end{itemize}
      Names may be multi-word (snake\_case). They should NOT be abstract role labels like ``agent" or
      ``instrument" — that's what the `role` field is for.
    \item A `role`: a single label from this closed vocabulary, describing the structural function the entity
      plays in its own domain:\{ROLE\_VOCAB\}. Pick the closest fit. Do not invent new role labels.
  \end{enumerate}

  CRITICAL: role labels must describe what the entity does in ITS OWN DOMAIN. Do NOT choose role labels to
  make the two domains' role lists look the same. If only the target domain has a clear `agent` and the
  source domain doesn't, that's fine — the asymmetry is informative. Forcing parallel role structures across
  domains defeats the purpose of the downstream structural-alignment scoring. Treat the two domains as
  independent role-typing exercises.

  YOUR OUTPUT WILL BE USED BY: Stage 3 will see only the `name` fields (not the roles) when extracting
  relations, so that its reasoning is grounded in concrete image content rather than abstract structure.

  Stage 4 will use the `role` fields to compute structural similarity between edges across domains.

  NODE RULES:
  \begin{itemize}
    \item 4–7 nodes per domain.
    \item Names: concrete, image-grounded, snake\_case, multi-word allowed. Be specific. Prefer
      ``construction\_worker" over ``agent\_figure"; prefer ``rusty\_nail" over ``metal\_object".
    \item Roles: from the closed vocabulary above, one per node.
    \item Names should be distinct within a domain. Roles MAY repeat within a domain (two `theme` nodes is
      fine if both genuinely play that structural part).
  \end{itemize}

  If a domain was identified as ``none" in Stage 1, return an empty list for that domain.
\end{promptbox}

\newpage

\begin{promptbox}
  {Stage 3: Relation Extraction}
  {prompt:relation-extraction}
  STAGE 3 of dual-domain graph extraction from a visual metaphor.

  YOU ARE GIVEN:
  \begin{itemize}
    \item the same image as Stage 1 and Stage 2
    \item the target and source domain names from Stage 1
    \item the per-domain node names from Stage 2 (CONCRETE names only, with role labels intentionally hidden)
  \end{itemize}

  YOUR TASK: for each domain, list the directed relations between its nodes. A relation is a triple (head,
  predicate, tail) where:
  \begin{itemize}
    \item head and tail are concrete node names from the Stage 2 list for that domain
    \item predicate captures a structural relationship that holds between the nodes within that domain
  \end{itemize}

  YOUR OUTPUT WILL BE USED BY: The categorical pullback algorithm, which compares the two graphs by finding a
  structure-preserving correspondence (between target\_nodes and source\_nodes) such that target\_edges align
  with source\_edges.

  Important: when you write a relation, focus on what is actually happening in the depicted image (for the
  source domain) or what genuinely holds in the abstract concept (for the target domain). Do NOT try to make
  the two domains' relations mirror each other — the pullback step will discover any genuine parallels on its own.

  RELATION RULES:
  \begin{itemize}
    \item Closed vocabulary ONLY: \{RELATION\_VOCAB\}. Pick the closest match if nothing is exact; do not
      invent new predicates.
    \item 3–7 edges per domain. More than 7 risks adding unsupported edges.
    \item Both head and tail MUST be exact matches to names in the node list for that domain.
    \item Multiple edges between the same pair are allowed (different predicates).
    \item Do NOT add edges between target nodes and source nodes — each graph is intra-domain only.
      Cross-domain mapping happens in the pullback step, not here.
    \item Assess each domain independently based on the image and your reasoning, and define the relations
      between the nodes that are most salient for each independent domain. Do not coordinate across domains.
  \end{itemize}

  If a domain has an empty node list, return an empty edge list for it.
\end{promptbox}

\newpage

\subsection{Iterative Refinement Prompt}

\begin{promptbox}
  {Refinement with Feedback Prompt}
  {prompt:feedback}
  You are refining the prompt given to a text-to-image model so that the image it produces is a STRONGER
  visual metaphor for a given idea.

  A visual metaphor depicts an abstract TARGET concept by portraying it through a concrete SOURCE concept,
  such that the relationships among the source elements mirror the relationships in the target idea.

  Here is the previous prompt you wrote and a diagnosis of how the image it produced fell short:

  PREVIOUS PROMPT: \{previous\_elaboration\}

  DIAGNOSIS: \{feedback\}

  Rewrite the prompt to fix the problems above. Specifically:
  \begin{itemize}
    \item Make BOTH the intended source and target concepts unmistakably present and recognisable in a single scene.
    \item Depict the relationships listed as missing, so the structure of the metaphor is visible, not just
      its surface objects.
    \item Keep the parts of the previous prompt that already worked; change only what the diagnosis flags.
  \end{itemize}

  Strict output rules:

  \begin{itemize}
    \item Output ONLY the new visual prompt itself.
    \item No preamble, no analysis, no headers, no markdown, no surrounding quotes, no commentary.
    \item Keep it under \{MAX\_WORDS\} words (~75 CLIP tokens).
    \item Write it as a single paragraph in natural prose.
  \end{itemize}
\end{promptbox}

\section{Can LLMs Extract and Populate Domain Graphs?}
\label{appendix:scar}

To ensure that \methodname's pullback score calculation pipeline (see \Cref{fig:extraction-pipeline}) is
feasible, we first validated whether LLMs are able to complete steps (2) and (3), the actual extraction of a
domain graph, for a specific domain. To test this behaviour, the SCAR benchmark \citep{yuan_beneath_2023} was
used, which contains analogous domains, their entities, and explanations of the relations between them.

Evaluating whether a given domain graph is valid or not is a difficult task. Therefore, we decided to follow
a meta-evaluation similar to the one presented by \cite{arora_healthbench_2025}, in which a human-annotated
binary response is used to test the accuracy of a given LLM-as-a-judge. We replicated this by selecting a
random sample of 100 domains from the SCAR dataset and manually annotating these. Then, we altered half of
the samples to produce incorrect domain graphs by altering or removing edges. Using this dataset, we assessed
whether an LLM can act as a judge that can distinguish between valid and invalid domain graphs. Specifically,
Claude Opus 4.6 \citep{anthropic_claude_2026} was used as the judge LLM and achieved an accuracy of 88\%,
with an F1 score of 0.867. Furthermore, out of the 50 invalid domain graphs, only one was found by Claude
Opus to be valid. Overall, Claude Opus 4.6 significantly outperforms the random baseline and errs
conservatively — misclassifying valid graphs as invalid rather than the reverse — making it a suitable judge
for this evaluation context.

Having established Claude Opus 4.6 as the judge, we then prompted Qwen 3.5 \citep{qwen_qwen_2026}, Gemma 4
\citep{gemma_gemma_2026}, and Claude Haiku 4.5 \citep{anthropic_claude_2025} to extract domain graphs for 100
randomly selected domains from the SCAR dataset, each with and without reasoning enabled. The results are
shown in \Cref{tab:benchmarking-domain-graph-results}. Gemma 4 with reasoning enabled and Claude Haiku 4.5
with extended thinking performed the best with high accuracies above 80\%. The results indicate that the
quality of domain graph extraction does differ significantly between different models, but that there is the
capability to extract valid domain graphs for various models, especially with reasoning modules enabled.

\begin{table}[H]
  \centering
  \small
  \caption{Results for the domain graph extraction task on 100 samples of the SCAR dataset
  \citep{yuan_beneath_2023}, as judged by Claude Opus 4.6.}
  \label{tab:benchmarking-domain-graph-results}
  \begin{tabular}{lc}
    \toprule
    \textbf{Model} & \textbf{Correctly generated graphs} \\
    \midrule
    Qwen 3.5         & 40\% \\
    Qwen 3.5 with reasoning & 76\% \\
    Gemma 4 & 78\% \\
    Gemma 4 with reasoning & 86\% \\
    Claude Haiku 4.5 & 56\% \\
    Claude Haiku 4.5 with extended thinking & 84\% \\
    \bottomrule
  \end{tabular}
\end{table}

\section{Can the Pullback Score Provide Useful Signals in Analogical Reasoning?}
\label{appendix:analobench}

This appendix provides extended analysis supporting the AnaloBench validation of the pullback score
summarised in \Cref{subsec:pullback-validation}. AnaloBench \citep{ye_analobench_2024} is a dataset of
stories of various lengths, some of which are analogous. Given a narrative, the task (T1) is to correctly
identify the analogous story from four given options \citep{ye_analobench_2024}. After determining that LLMs
are able to extract valid domain graphs (see \Cref{appendix:scar}), we used this task to configure the
pullback score algorithm and its extraction prompts, and to validate whether the resulting score provides a
useful signal for analogical reasoning, comparing a zero-shot baseline (GPT-4.1 Mini directly selecting the
analogous story) against selecting the candidate with the highest pullback score.

\begin{figure}[H]
  \centering
  \includegraphics[width=0.8\linewidth]{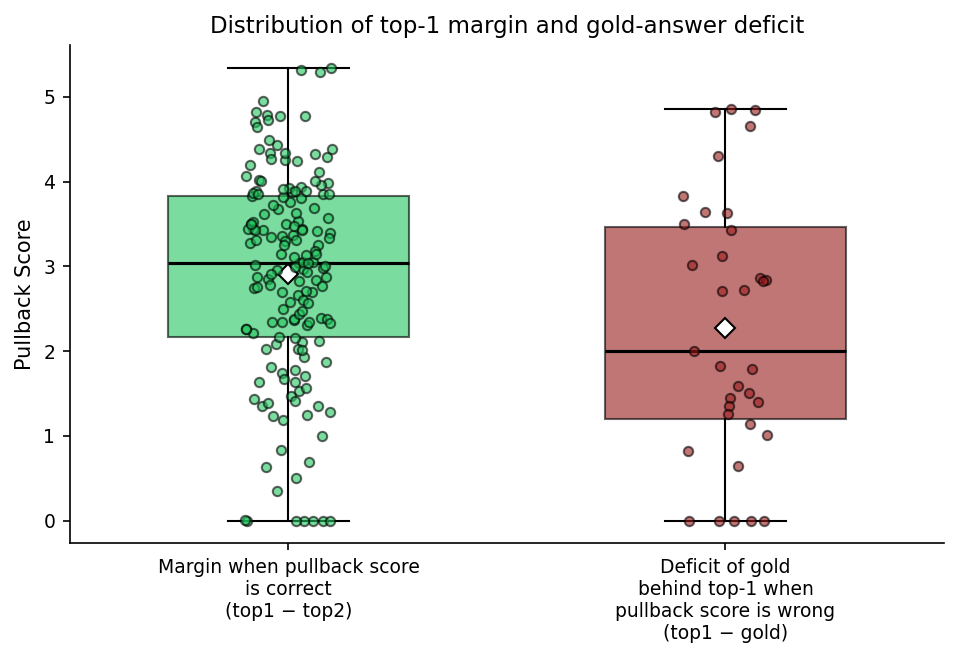}
  \caption{Confidence margins of the pullback score on the AnaloBench task: for correct predictions, the gap
    between the highest- and second-highest-scoring candidates; for incorrect predictions, the gap between the
    highest-scoring (wrong) candidate and the score of the true answer. Correct predictions show consistently
    larger margins than incorrect ones, indicating that the pullback score expresses greater certainty
  precisely when it identifies the correct analogy.}
  \label{fig:analobench_margins}
\end{figure}

We analysed the confidence of the pullback approach by examining the score margins between candidate answers.
Figure~\ref{fig:analobench_margins} shows the distribution of the difference between the highest-ranked and
second-highest-ranked pullback scores for correctly answered examples, alongside the difference between the
highest-ranked score and the score assigned to the true answer for incorrectly answered examples. The margins
for correct predictions were generally larger than those for incorrect predictions, suggesting that the
pullback score expresses greater certainty when it identifies the correct analogy.

Qualitative inspection of the cases where the zero-shot approach succeeded while the pullback score approach
failed suggested a potential bias towards denser and more highly connected graphs. In several of these cases,
the pullback score assigned the highest score to an incorrect candidate whose extracted graph contained a
larger number of (unmatched) relations than the graph corresponding to the correct answer. This observation
highlights a potential limitation of the current scoring mechanism and suggests a direction for future refinement.

\section{Algorithm Description}
\label{appendix:algorithms}

This subroutine implements the matching step invoked by the pullback score computation introduced in
\Cref{sec:methodology} (\Cref{alg:pullback}). \textsc{GreedyPullback} is a single-pass greedy heuristic:
candidate edge pairs are sorted by similarity and accepted in order subject to the node-consistency check.
\begin{algorithm}[H]
  \caption{Greedy Semantic Pullback Matching}
  \label{alg:greedy}
  \footnotesize
  \begin{algorithmic}[1]

    \Require Source edge set \(E^S = \{(u_i, r_i, v_i)\}_{i=1}^{m}\), target
    edge set \(E^T = \{(u'_j, r'_j, v'_j)\}_{j=1}^{n}\), precomputed embeddings
    \(\hat{E}^S \in \mathbb{R}^{m \times d}\),
    \(\hat{E}^T \in \mathbb{R}^{n \times d}\), matching threshold \(\theta\)
    \Ensure Pullback score \(\psi\), matched pairs \(\mathcal{P}\), coverage \(\rho\)
    \Procedure{GreedyPullback}{\ensuremath{E^S, E^T, \hat{E}^S, \hat{E}^T, \theta}}
    \State \(\Phi \gets \hat{E}^S (\hat{E}^T)^\top\)
    \Comment{Cosine similarity matrix \(\in \mathbb{R}^{m \times n}\)}
    \State \(M \gets \emptyset\)
    \Comment{Node mapping: source node \(\to\) target node}
    \State \(U_S \gets \emptyset\), \(U_T \gets \emptyset\)
    \Comment{Sets of already-matched edge indices}
    \State \(\mathcal{P} \gets \emptyset\)
    \Comment{Accepted edge pairs}
    \State \(C \gets \{(\Phi_{ij}, i, j) \mid \Phi_{ij} \geq \theta\}\) sorted in
    descending order of \(\Phi_{ij}\)
    \For{\textbf{each} \((\varphi, i, j) \in C\)}
    \If{\(i \in U_S\) \textbf{or} \(j \in U_T\)}
    \State \textbf{continue}
    \Comment{Each edge may be matched at most once}
    \EndIf
    \State Let \(e^S_i = (u_i, r_i, v_i)\) and \(e^T_j = (u'_j, r'_j, v'_j)\)
    \State \(\Delta \gets \{u_i \mapsto u'_j,\; v_i \mapsto v'_j\}\)
    \Comment{Proposed node assignments for this pair}
    \If{\(\exists\, (a \mapsto b) \in M\) such that \(a \in \operatorname{dom}(\Delta)\) and
    \(\Delta(a) \neq b\)}
    \State \textbf{continue}
    \Comment{Reject: violates global node consistency (pullback
    condition)}
    \EndIf
    \State \(M \gets M \cup \Delta\)
    \State \(U_S \gets U_S \cup \{i\}\),\quad \(U_T \gets U_T \cup \{j\}\)
    \State \(\mathcal{P} \gets \mathcal{P} \cup \{(i, j, \varphi)\}\)
    \EndFor
    \State \(\sigma \gets \sum_{(i, j, \varphi) \in \mathcal{P}} \varphi\)
    \Comment{Total matched similarity, \Cref{eq:pullback-score}}
    \State \(\rho \gets |\mathcal{P}| \,/\, |E^T|\)
    \Comment{Coverage: fraction of target edges matched}
    \State \(\ell \gets \textsc{LongestSimplePath}\big(G^T[\mathcal{P}]\big)\)
    \Comment{\(G^T[\mathcal{P}]\): subgraph of \(G^T\) induced by matched edges; \(\ell\) counted in edges}
    \State \(\psi \gets \sigma + 0.5\,\ell\)
    \Comment{Pullback score, \Cref{eq:pullback-score}}
    \State \Return \(\psi\), \(\mathcal{P}\), \(\rho\)
    \EndProcedure

  \end{algorithmic}
\end{algorithm}

\newpage

\section{User Study Instructions}
\label{appendix:user-study}

\begin{figure}[H]
  \centering
  \includegraphics[width=1\linewidth]{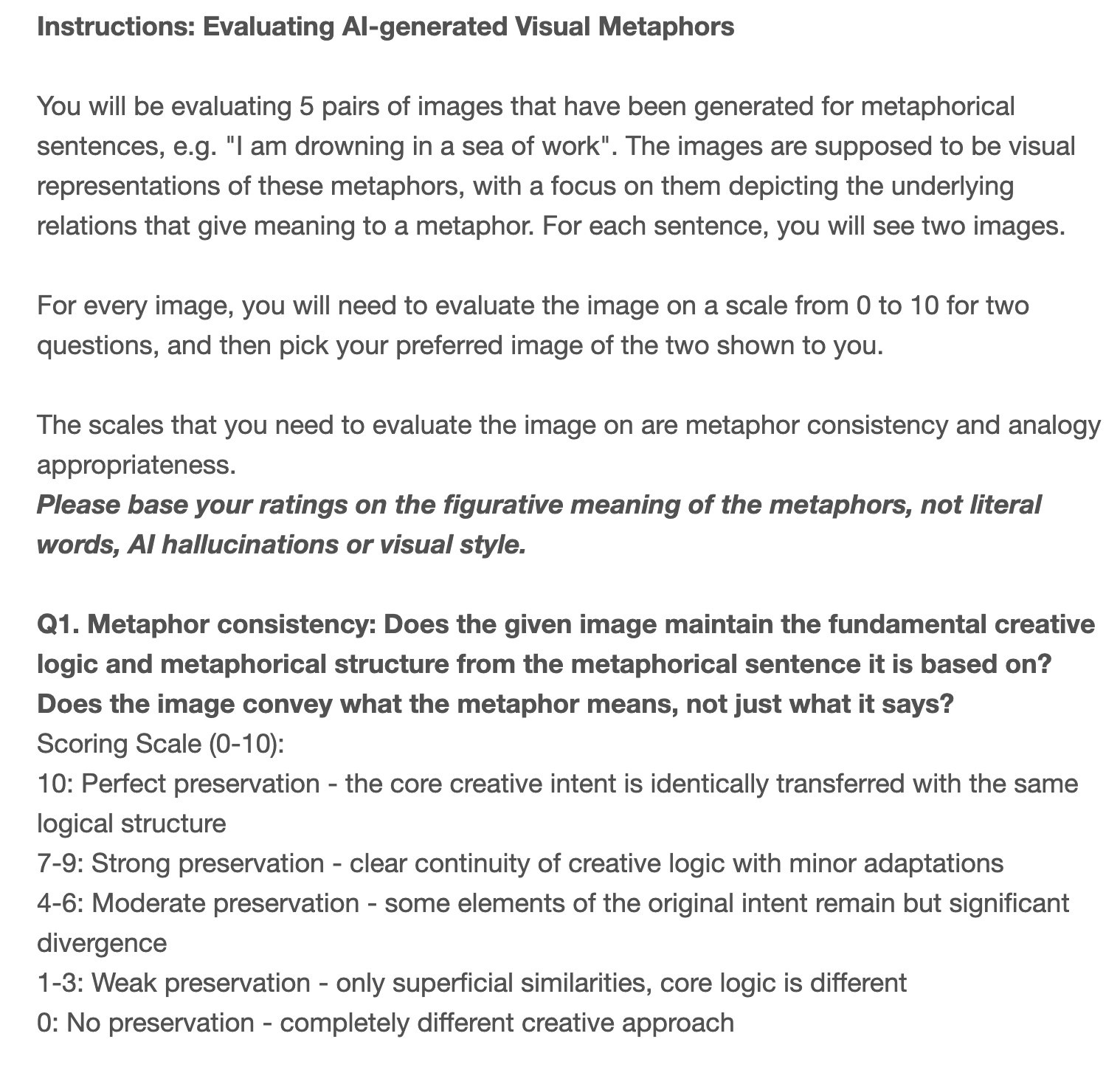}
  \caption{Part 1 of instructions provided to participants in the human evaluation.}
  \label{fig:user-study-1}
\end{figure}

\newpage

\begin{figure}[H]
  \centering
  \includegraphics[width=1\linewidth]{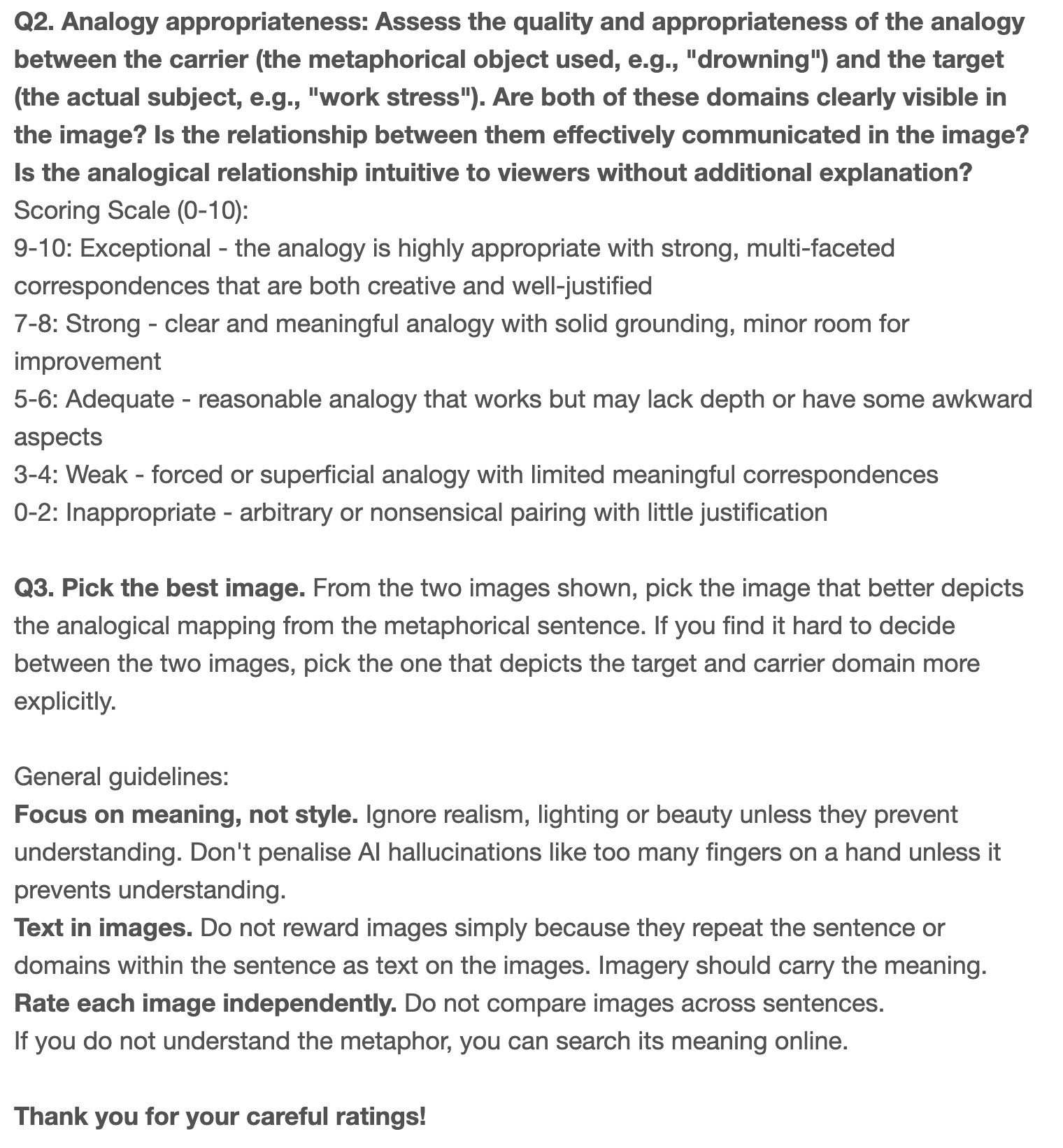}
  \caption{Part 2 of instructions provided to participants in the human evaluation.}
  \label{fig:user-study-2}
\end{figure}

\newpage

\section{Zero-Shot Generation Examples}
\label{appendix:zero-shot-examples}

This appendix provides qualitative examples of the visual elaborations and generated images.

\setlength{\eximgwidth}{0.28\textwidth}

\begin{table}[H]
  \centering
  \small
  \setlength{\tabcolsep}{8pt}
  \renewcommand{\arraystretch}{1.2}
  \caption{%
    Selected zero-shot generated visual metaphors.
    Each row shows the source metaphor, the model's visual elaboration, and the resulting generated image.
  }
  \label{tab:appendix-zero-shot}
  \begin{tabular}{%
      >{\RaggedRight\arraybackslash}m{0.15\textwidth}   % Metaphor
      >{\RaggedRight\arraybackslash}m{0.47\textwidth}   % Visual Elaboration
      >{\centering\arraybackslash}m{\eximgwidth}         % Image
    }
    \toprule
    \textbf{Metaphor} & \textbf{Visual Elaboration} & \textbf{Generated Image (GPT Image 1.5)} \\
    \midrule

    Love is a double edged sword. &
    A gleaming ornate sword suspended in mid-air, its blade splitting into two distinct halves—one radiating
    warm golden light and blooming roses, the other crackling with cold blue electricity and thorns. The
    background fades between serene twilight and stormy darkness. Fine mist swirls around the weapon,
    capturing the duality of beauty and pain. &
    \includegraphics[width=\linewidth]{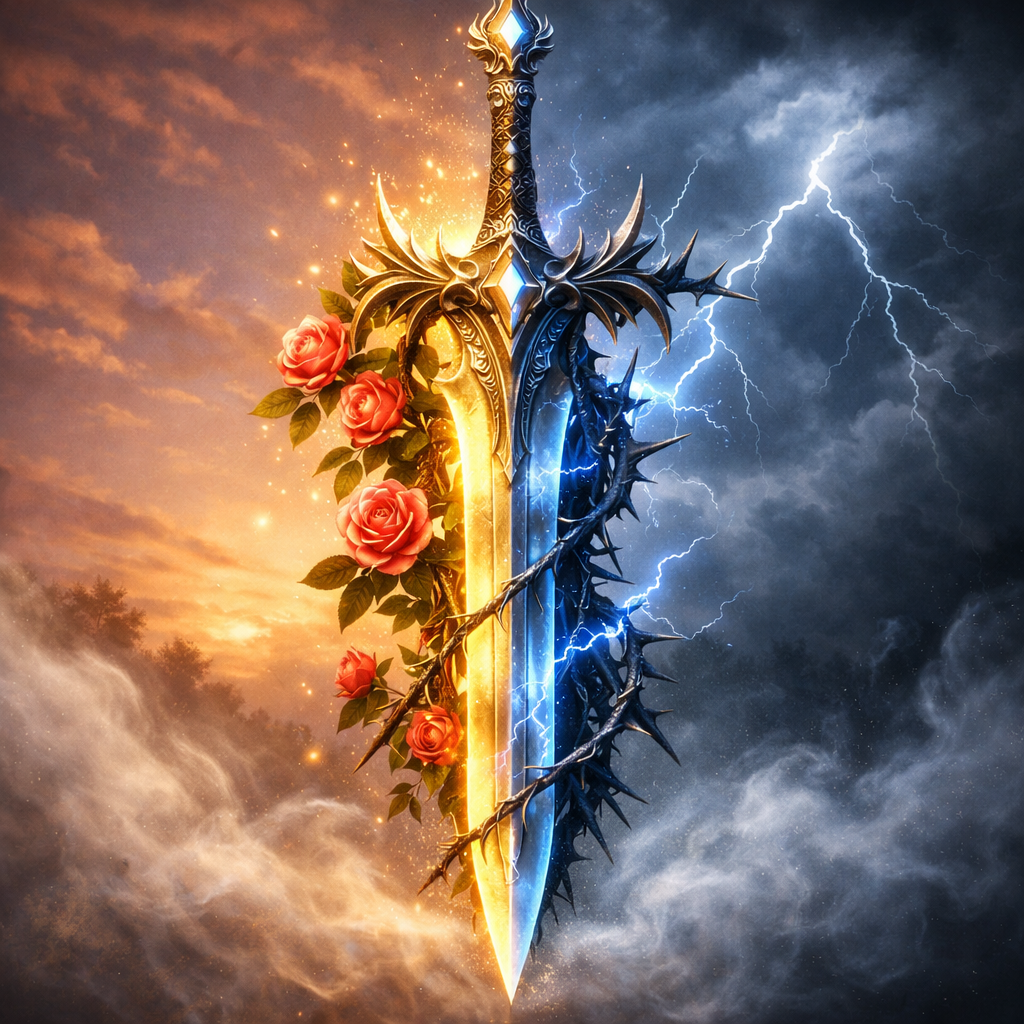} \\

    \midrule

    A sweet tooth is a predator chasing down your smile. &
    A sleek, shadowy predator with candy-colored fur stalks through a dreamlike landscape toward a luminous,
    golden smile floating in the distance. The creature's eyes glow with hunger as it prowls closer, leaving
    a trail of melting sweets and broken teeth. Soft, surreal lighting contrasts the predator's dark
    silhouette against pastel clouds and candy-striped terrain. &
    \includegraphics[width=\linewidth]{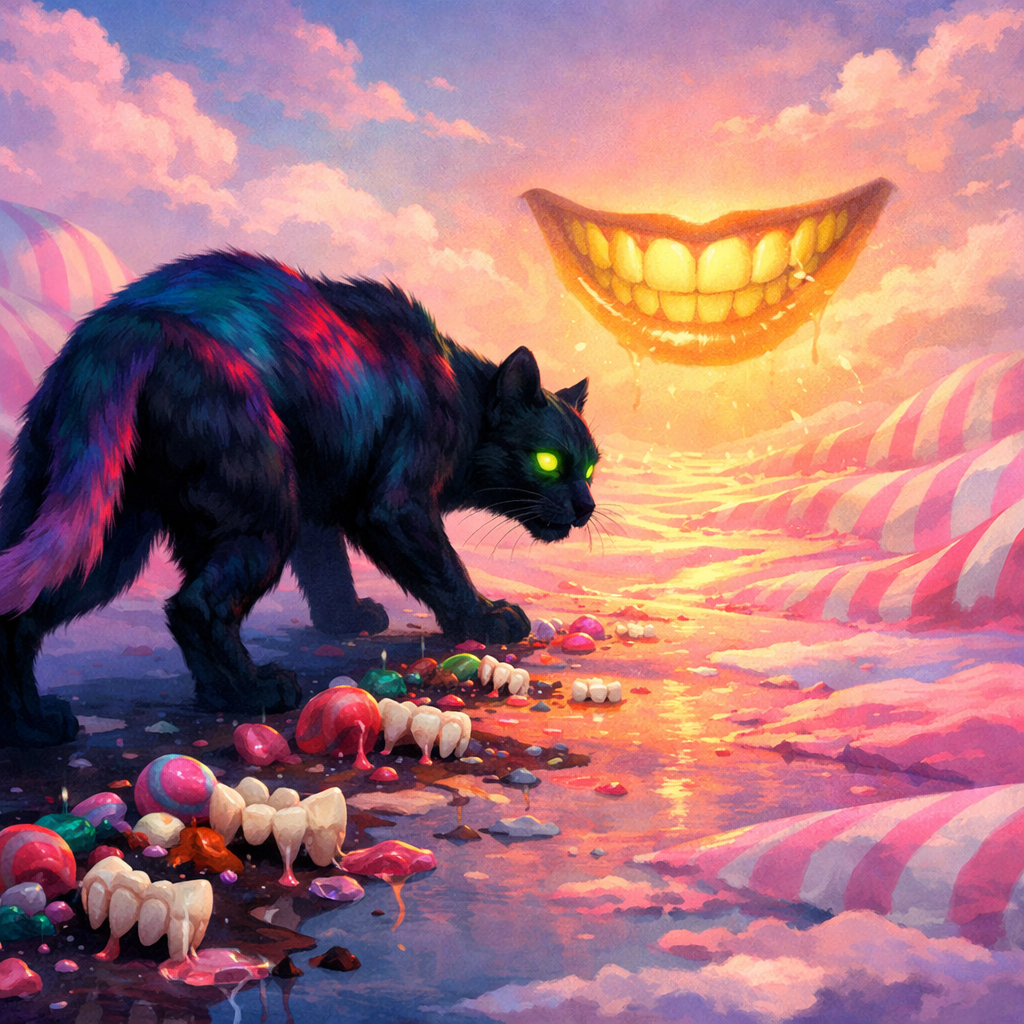} \\

    \midrule

    The planet is a sinking ship. &
    A massive Earth sphere tilts precariously in turbulent waters, its continents cracking and flooding.
    Desperate figures cling to the edges as waves crash over the surface. Smoke rises from fissures. The sky
    darkens ominously. Lifeboats drift empty nearby, unreachable. The horizon swallows everything in murky
    depths, evoking apocalyptic urgency and collective doom. &
    \includegraphics[width=\linewidth]{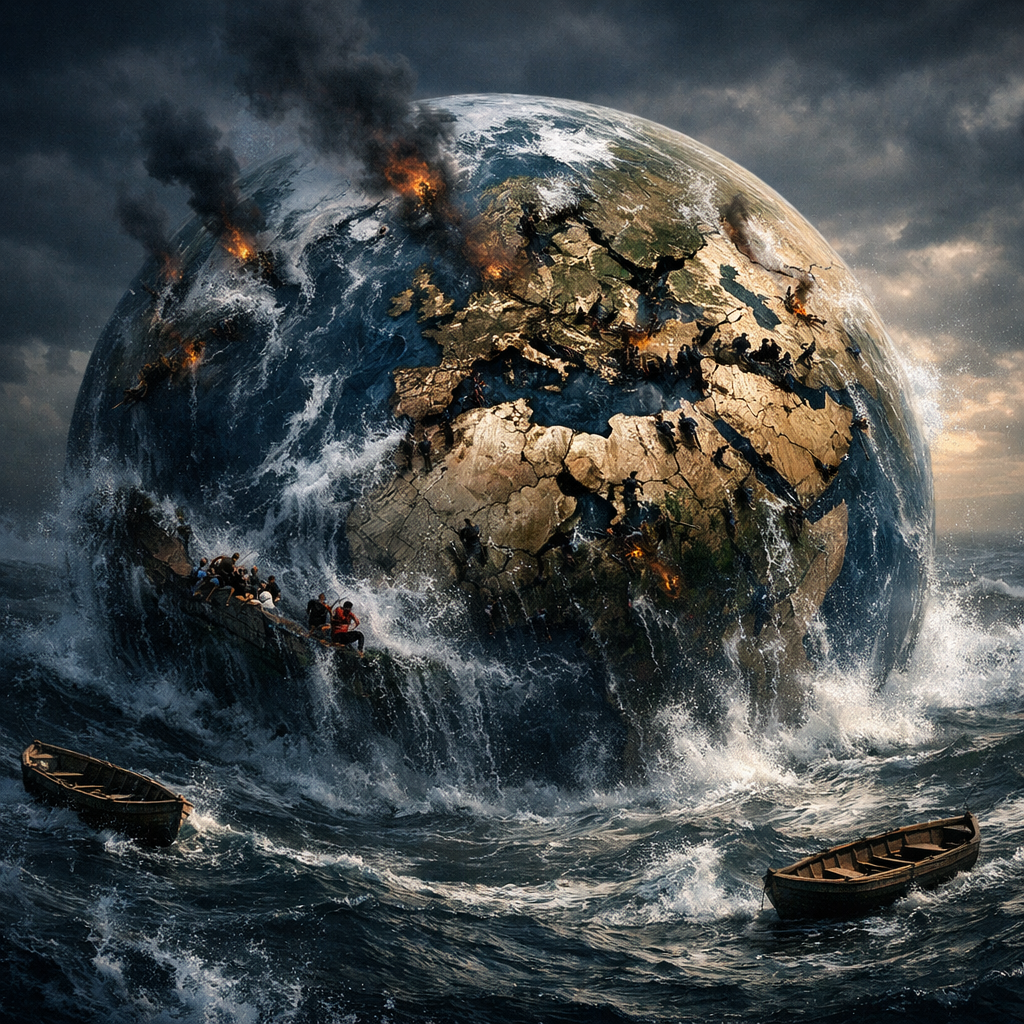} \\

    \midrule

    Social media is a hamster wheel. &
    An exhausted figure endlessly running inside a massive transparent hamster wheel, surrounded by glowing
    smartphone screens and notification badges. The wheel spins relentlessly in a dimly lit room, casting
    repetitive shadows. Despite constant motion, the scenery never changes. Scattered digital clutter
    accumulates around the base as the person runs faster, trapped in an endless cycle of movement without progress. &
    \includegraphics[width=\linewidth]{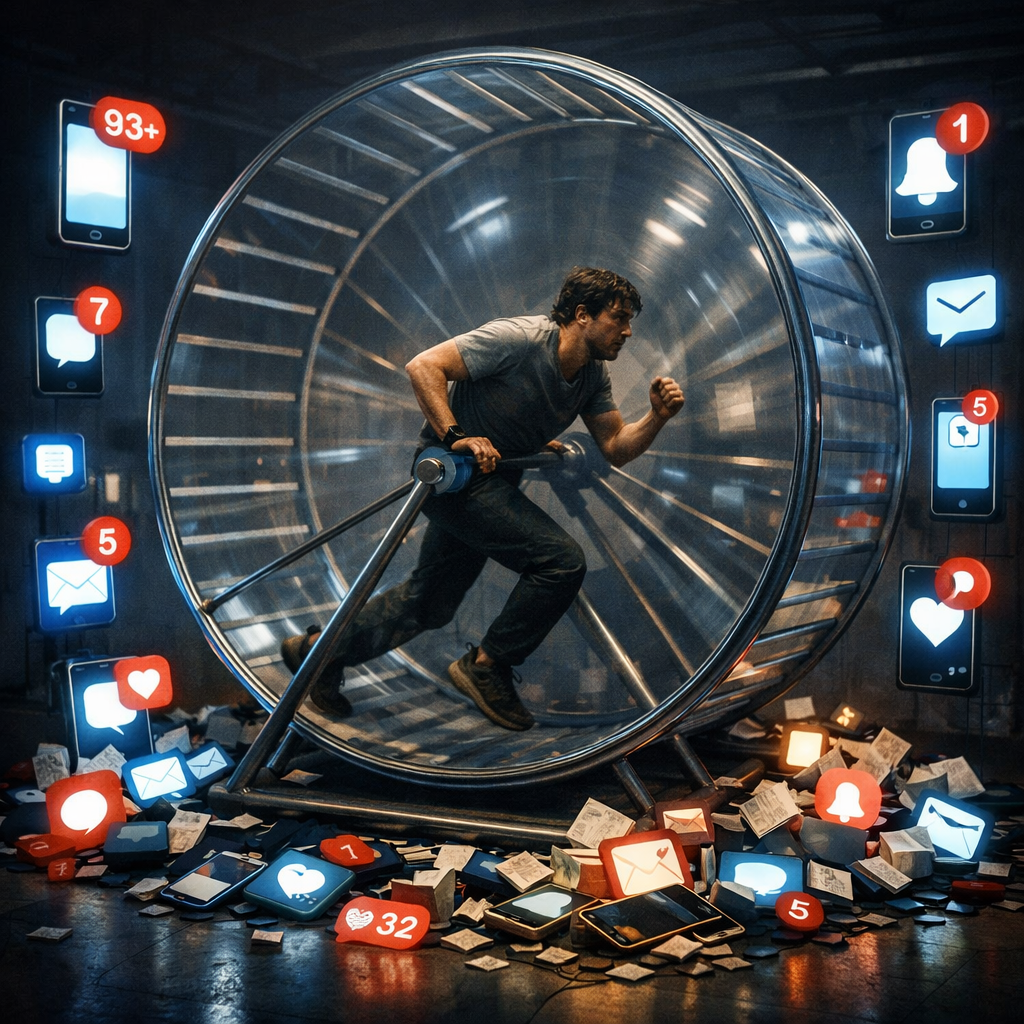} \\

    \bottomrule
  \end{tabular}
\end{table}

\newpage

\section{Extended Zero-Shot Results}
\label{appendix:zero-shot-quant}

This appendix reports extended results for zero-shot visual metaphors generated with Flux.2 Klein
\citep{black_forest_labs_black-forest-labsflux2-klein-4b_2026}, GPT-Image 1.5
\citep{openai_gpt_2025}, and Stable Image Core v1.1 \citep{stability_ai_stability_2024}.

\paragraph{Metrics.}
We evaluate the generated images using domain fidelity, CLIP retrieval accuracy, and the pullback score.
Domain fidelity measures the cosine similarity between the gold-standard source and target domains and the
domains extracted from the generated image. CLIP retrieval accuracy
\citep{radford_learning_2021,hessel_clipscore_2021} measures consistency with the input metaphor by reporting
how often the corresponding text or image is retrieved at rank 1 or within the top 5, following
retrieval-based text-to-image evaluation protocols \citep{cho_dall-eval_2023}. The pullback score measures
relational alignment between the domain graphs extracted from each image.

\paragraph{Generated images rarely depict both intended domains recognisably.}
As shown in \Cref{tab:domain-fidelity}, fewer than 10\% of images recognisably depict both input domains
under every condition, with GPT-Image zero-shot performing best at 9.4\%. Either the source or target domain
is recognisable more often, reaching 59.2\% in the same condition. Although domain extraction admits multiple
valid interpretations, these results indicate that simultaneously communicating both sides of a metaphor
remains difficult for current image generators.

\begin{table}[H]
  \centering
  \small
  \caption{Domain fidelity: percentage of metaphors for which the generated image recognisably depicts both
  or either input domain. Fidelity to both domains remains below 10\% across all conditions.}
  \label{tab:domain-fidelity}
  \begin{tabular}{lcc}
    \toprule
    \textbf{Model} & \textbf{Both Domains Match} & \textbf{Either Domain Matches} \\
    \midrule
    GPT-Image Zero-Shot          & \textbf{9.4\%} & \textbf{59.2\%} \\
    GPT-Image Chain-of-Thought   & 6.5\% & 55.7\% \\
    Flux-Klein Zero-Shot         & 5.9\% & 54.2\% \\
    Stable-Core Chain-of-Thought & 4.8\% & 44.0\% \\
    Stable-Core Zero-Shot        & 4.4\% & 52.0\% \\
    Flux-Klein Chain-of-Thought  & 4.2\% & 52.9\% \\
    \bottomrule
  \end{tabular}
\end{table}

\paragraph{GPT-Image zero-shot best preserves the input metaphor.}
CLIP retrieval produces a similar ranking. GPT-Image zero-shot achieves the highest retrieval accuracy in
both directions, including 84\% image-to-text Top-1 accuracy, compared with 59\% for Stable-Core under
chain-of-thought prompting (\Cref{tab:clip-fidelity}). Zero-shot prompting outperforms chain-of-thought
prompting for all three generators, suggesting that the additional elaboration introduced by chain-of-thought
prompting often causes the image to drift from the original metaphor. We therefore use GPT-Image zero-shot as
the initial generation condition in the refinement experiments.

\begin{table}[H]
  \centering
  \caption{CLIP retrieval accuracy between generated visual metaphors and their input texts. Higher Top-1 and
  Top-5 accuracy and lower mean rank indicate better retrieval.}
  \label{tab:clip-fidelity}
  \small
  \setlength{\tabcolsep}{4pt}
  \begin{tabular}{lccc|ccc}
    \toprule
    &
    \multicolumn{3}{c|}{\textbf{Image \(\rightarrow\) Text}} &
    \multicolumn{3}{c}{\textbf{Text \(\rightarrow\) Image}} \\
    \cmidrule(r){2-4}
    \cmidrule(l){5-7}
    \textbf{Model}
    & \textbf{Top-1 (\%)}
    & \textbf{Top-5 (\%)}
    & \textbf{Mean Rank}
    & \textbf{Top-1 (\%)}
    & \textbf{Top-5 (\%)}
    & \textbf{Mean Rank} \\
    \midrule
    GPT-Image Zero      & \textbf{84} & \textbf{96} & \textbf{1.57}
    & \textbf{82} & \textbf{96} & \textbf{2.54} \\
    GPT-Image CoT       & 76 & 91 & 3.19 & 72 & 90 & 3.57 \\
    Flux-Klein Zero     & 76 & 92 & 2.60 & 67 & 87 & 4.50 \\
    Stable-Core Zero    & 69 & 87 & 4.22 & 61 & 82 & 6.35 \\
    Flux-Klein CoT      & 65 & 86 & 4.27 & 59 & 81 & 5.33 \\
    Stable-Core CoT     & 59 & 81 & 7.08 & 54 & 77 & 7.84 \\
    \bottomrule
  \end{tabular}
\end{table}

\paragraph{Relational structure can remain coherent even when the intended domains are not.}
Pullback scores exhibit less variation across conditions than either domain fidelity or CLIP retrieval.
Scores are concentrated between 1 and 4, and comparisons against the human condition find no significant
difference for any generator except Stable-Core zero-shot (\(p_{\text{bonf}}=0.040\)). Human-created metaphors
nevertheless produce non-zero scores most consistently, at 94.7\%, compared with 77.0--93.8\% across the
generated conditions.

\begin{figure}[H]
  \centering
  \includegraphics[width=1\linewidth]{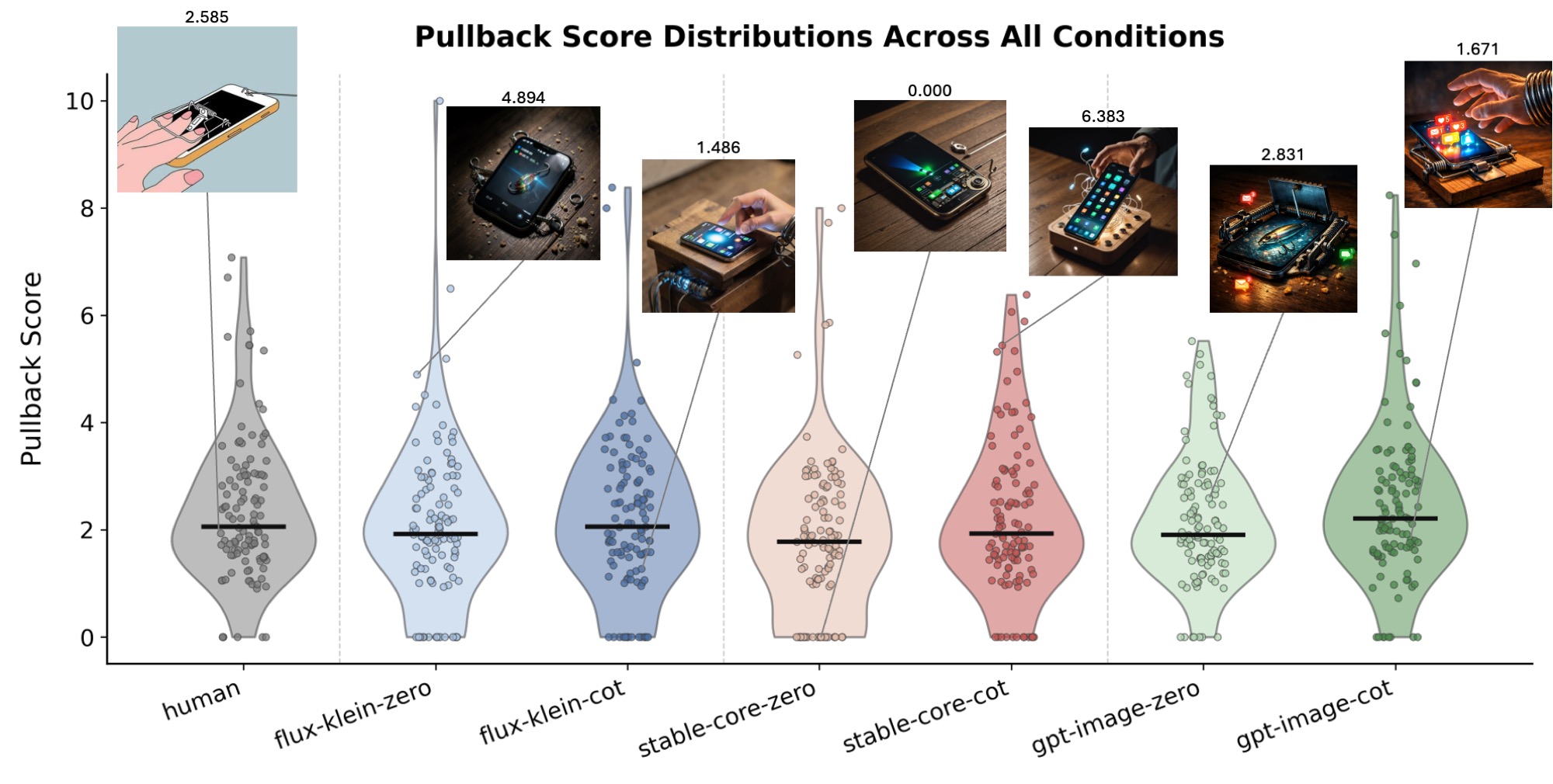}
  \caption{Distribution of pullback scores across six generation conditions and the human baseline. Annotated
  points show results for the metaphor ``Your phone is a mouse trap''.\protect\footnotemark}
  \label{fig:violin-plot-pullback-scores}
\end{figure}
\footnotetext{Source of the human-created image for the metaphor ``Your phone is a mouse trap'': Pinterest
(original creator unknown), \url{https://www.pinterest.com/pin/788833691001985793/}. Accessed 6 July 2026.}

This result does not imply that generated images preserve the intended analogy as faithfully as human-created
metaphors. The pullback score evaluates alignment between the domains extracted from an image, regardless of
whether these are the intended domains. Taken together, the metrics therefore reveal a distinction between
structural coherence and conceptual fidelity: generated images often contain a coherent relation between two
depicted domains, but those domains frequently differ from the source and target specified by the input
metaphor. \methodname addresses this gap by steering generation towards the particular relational structure
intended by the input.

\begin{table}[H]
  \centering
  \caption{Correlations between metaphor properties and CLIP retrieval accuracy. Sample sizes (\(n\)) are below
    the full 250 metaphors in some conditions because a subset of generations was blocked by the image
    generator's content moderation filters and could therefore not be scored. Significance is assessed using a
    Bonferroni-corrected threshold of \(\alpha=0.05/6\approx0.0083\) within each analysis. Significant \(p\)-values
  are shown in bold.}
  \label{tab:difficulty-correlations}
  \small
  \setlength{\tabcolsep}{4.5pt}
  \begin{tabular}{llccccc}
    \toprule
    \textbf{Analysis} & \textbf{Condition} & \(n\)
    & \textbf{Pearson \(r\)} & \(p\)
    & \textbf{Spearman \(\rho\)} & \(p\) \\
    \midrule
    \multirow{6}{*}{Concreteness}
    & Flux-Klein Zero   & 231 & \(+0.230\) & \(\mathbf{0.0004}\) & \(+0.217\) & \(\mathbf{0.0009}\) \\
    & Stable-Core Zero  & 243 & \(+0.186\) & \(\mathbf{0.0035}\) & \(+0.171\) & \(\mathbf{0.0074}\) \\
    & GPT-Image Zero    & 238 & \(+0.231\) & \(\mathbf{0.0003}\) & \(+0.243\) & \(\mathbf{0.0002}\) \\
    & Flux-Klein CoT    & 231 & \(+0.276\) & \(\mathbf{<0.0001}\) & \(+0.278\) & \(\mathbf{<0.0001}\) \\
    & Stable-Core CoT   & 243 & \(+0.208\) & \(\mathbf{0.0011}\) & \(+0.191\) & \(\mathbf{0.0029}\) \\
    & GPT-Image CoT     & 239 & \(+0.238\) & \(\mathbf{0.0002}\) & \(+0.259\) & \(\mathbf{0.0001}\) \\
    \midrule
    \multirow{6}{*}{Nearness}
    & Flux-Klein Zero   & 238 & \(-0.064\) & \(0.3225\) & \(-0.066\) & \(0.3136\) \\
    & Stable-Core Zero  & 250 & \(-0.032\) & \(0.6145\) & \(-0.051\) & \(0.4242\) \\
    & GPT-Image Zero    & 245 & \(-0.131\) & \(0.0407\) & \(-0.136\) & \(0.0333\) \\
    & Flux-Klein CoT    & 238 & \(-0.065\) & \(0.3149\) & \(-0.074\) & \(0.2551\) \\
    & Stable-Core CoT   & 250 & \(-0.002\) & \(0.9805\) & \(-0.020\) & \(0.7528\) \\
    & GPT-Image CoT     & 246 & \(-0.042\) & \(0.5157\) & \(-0.050\) & \(0.4367\) \\
    \bottomrule
  \end{tabular}
\end{table}

\paragraph{What makes a metaphor harder to depict?}
We additionally examine whether CLIP retrieval accuracy varies with two properties of the input metaphor:
between-domain nearness and domain concreteness. Nearness is measured as the cosine similarity between the
gold-standard source and target domains, with higher values indicating more similar domains. This analysis
tests whether the advantage of near over far analogies previously observed in language-model reasoning
\citep{khojasteh_enhancing_2026} also appears in visual metaphor generation.

Concreteness is additionally measured using the ratings of \citet{brysbaert_concreteness_2014}. Because a
metaphor is constrained by its harder-to-visualise domain, we define its concreteness as the lower rating of
its source and target domains, where ratings range from 1 for highly abstract concepts to 5 for highly
concrete concepts.

As shown in \Cref{tab:difficulty-correlations}, concreteness is positively associated with retrieval accuracy
in every condition, with small but consistent correlations (\(r=0.186\)--\(0.276\)). Metaphors are therefore
easier to depict faithfully when both domains refer to readily visualisable concepts. By contrast, nearness
shows no significant relationship with retrieval accuracy after correction for multiple comparisons. The
difficulty of visual metaphor generation thus appears to depend more on whether the constituent domains can
be rendered concretely than on their semantic similarity. Unlike language-based analogy selection, image
generation does not receive a clear advantage from greater surface overlap between the domains.

\section{Model Selection \& Hyperparameter Tuning}
\label{appendix:hyperparameters}

\paragraph{Refinement model selection.}
We conducted hyperparameter tuning experiments to select the most suitable model for generating the visual
elaborations and the feedback metrics based on the pullback score extraction. In particular, we evaluated
three state-of-the-art models: Gemma 4 31B, Claude Sonnet 4.6, and GPT 5.5.

\paragraph{Stopping criteria.}
Beyond model selection, we also wanted to investigate appropriate values for the stopping criteria (i.e. the
maximum number of refinement rounds, \(k\), and the convergence threshold, \(\varepsilon\)). In principle, the
refinement loop could continue until a target pullback score is reached. In practice, however, computational
and financial constraints make an unbounded number of iterations infeasible. Consequently, we treated both
\(k\) and \(\varepsilon\) as tunable hyperparameters. We considered \(k \in \{5,10,15\}\) and \(\varepsilon \in
\{0.05,0.1,0.2\}\).

\begin{table}[H]
  \centering
  \caption{Hyperparameter tuning results for different visual elaboration models and stopping criteria.
  Scores correspond to the average ratings assigned by the three independent VLM judges.}
  \label{tab:hyperparameter_tuning}
  \resizebox{\textwidth}{!}{
    \begin{tabular}{llccccc}
      \toprule
      \textbf{Model} & \(\boldsymbol{k}\) & \(\boldsymbol{\varepsilon}\) & \textbf{Mean Pullback} &
      \textbf{Metaphor Consistency} &
      \textbf{Analogy Appropriateness} &
      \textbf{Conceptual Integration} \\
      \midrule

      Gemma 4 31B & 5  & 0.05 & 2.8874 & 8.1534 & 8.6399 & 8.6667 \\
      Gemma 4 31B & 5  & 0.10 & 2.8100 & 8.0467 & 8.5933 & 8.6000 \\
      Gemma 4 31B & 5  & 0.20 & 2.8064 & 8.0467 & 8.5933 & 8.6134 \\
      Gemma 4 31B & 10 & 0.05 & 3.1357 & 7.9800 & 8.4800 & 8.6200 \\
      Gemma 4 31B & 10 & 0.10 & 3.0188 & 7.9000 & 8.4400 & 8.5667 \\
      Gemma 4 31B & 10 & 0.20 & 2.9070 & 7.9400 & 8.4533 & 8.5867 \\
      Gemma 4 31B & 15 & 0.05 & 3.1851 & 7.9800 & 8.5266 & 8.6334 \\
      Gemma 4 31B & 15 & 0.10 & 3.0365 & 7.9000 & 8.4866 & 8.5934 \\
      Gemma 4 31B & 15 & 0.20 & 2.9202 & 7.9267 & 8.5066 & 8.6267 \\
      \midrule

      Claude Sonnet 4.6 & 5  & 0.05 & 3.6607 & 8.1333 & 8.7266 & 8.7201 \\
      Claude Sonnet 4.6 & 5  & 0.10 & 3.6128 & 8.1866 & 8.7266 & \textbf{8.7334} \\
      Claude Sonnet 4.6 & 5  & 0.20 & 3.5388 & 8.1199 & 8.6866 & 8.6934 \\
      Claude Sonnet 4.6 & 10 & 0.05 & 4.1000 & 8.1599 & 8.7266 & 8.7201 \\
      Claude Sonnet 4.6 & 10 & 0.10 & 3.9788 & \textbf{8.2133} & \textbf{8.7333} & 8.7001 \\
      Claude Sonnet 4.6 & 10 & 0.20 & 3.8795 & 8.1599 & 8.6866 & 8.6734 \\
      Claude Sonnet 4.6 & 15 & 0.05 & \textbf{4.1079} & 8.1599 & 8.7000 & 8.7134 \\
      Claude Sonnet 4.6 & 15 & 0.10 & 3.9867 & \textbf{8.2133} & 8.7067 & 8.6934 \\
      Claude Sonnet 4.6 & 15 & 0.20 & 3.8874 & 8.1599 & 8.6600 & 8.6668 \\
      \midrule

      GPT-5.5 & 5  & 0.05 & 2.7944 & 7.7332 & 8.3916 & 8.0832 \\
      GPT-5.5 & 5  & 0.10 & 2.7231 & 7.6666 & 8.3083 & 8.1499 \\
      GPT-5.5 & 5  & 0.20 & 2.3086 & 7.6666 & 8.2250 & 8.1666 \\
      GPT-5.5 & 10 & 0.05 & 3.2636 & 7.1998 & 7.9416 & 7.5667 \\
      GPT-5.5 & 10 & 0.10 & 3.0258 & 7.1665 & 7.8916 & 7.6334 \\
      GPT-5.5 & 10 & 0.20 & 2.4495 & 7.6999 & 8.2750 & 8.1834 \\
      GPT-5.5 & 15 & 0.05 & 3.2636 & 7.1998 & 7.9416 & 7.5667 \\
      GPT-5.5 & 15 & 0.10 & 3.0258 & 7.1665 & 7.8916 & 7.6334 \\
      GPT-5.5 & 15 & 0.20 & 2.4495 & 7.6999 & 8.2750 & 8.1834 \\
      \bottomrule
    \end{tabular}
  }
\end{table}

\paragraph{Calibration evaluation.}
Performing human evaluations for every configuration was impractical, so a panel of VLMs were used as judges
instead. The VLM judge models used were Gemini 3.5 Flash, Claude Opus 4.8 and Llama-4 Maverick.\footnote{This
  calibration run used the three-judge panel in place at the time. The judging panel used for all subsequent
evaluations reported in this paper is the two-judge panel described in \Cref{subsec:metrics}.}

This analysis was performed on a randomly selected holdout set of 25 visual metaphors, which is disjoint from
the 200-metaphor evaluation set reported in \Cref{sec:results}. The results are shown in
\Cref{tab:hyperparameter_tuning}. Overall, Claude Sonnet 4.6 consistently outperformed Gemma 4 31B and
GPT-5.5 across all evaluation settings and was therefore selected as the visual elaboration generator and
feedback model.

The highest pullback score was achieved by Claude Sonnet 4.6 with \(k=15\) and \(\varepsilon=0.05\). However,
this configuration only marginally outperformed the \(k=10\), \(\varepsilon=0.05\) setting (4.1079 vs. 4.1000),
while requiring 50\% more refinement iterations. Since the additional gains in both pullback score and
VLM-based evaluation metrics were negligible across all configurations with \(k \geq 10\), we selected \(k=10\)
as a more computationally efficient operating point.

Although \(\varepsilon=0.10\) obtains marginally higher Metaphor Consistency and Analogy Appropriateness than
\(\varepsilon=0.05\) at \(k=10\) (8.2133 vs.\ 8.1599, and 8.7333 vs.\ 8.7266), these differences are small
relative to the noise expected from scoring only 25 holdout metaphors with the preliminary three-judge panel,
whereas \(\varepsilon=0.05\) achieves a larger, more decisive margin on Conceptual Integration (8.7201
vs.\ 8.7001) and on the mean pullback score (4.1000 vs.\ 3.9788). Since \(\varepsilon\) is defined as the
convergence threshold on the pullback score itself rather than on the downstream judge scores, we prioritised
this larger, directly targeted margin over marginal, likely noise-level differences on axes \(\varepsilon\) was
not selected to optimise.

\paragraph{Pullback matching threshold.}
The pullback score uses a cosine-similarity threshold, \(\theta\), to determine whether source- and
target-domain relation edges are sufficiently similar to be matched. We calibrated \(\theta\) on the pullback
scores of the 125 original, human-authored visual metaphor images themselves by varying \(\theta\) from 0.35 to
0.55 and measuring the resulting mean pullback score, the proportion of zero scores, and correlation with
CLIP image-text similarity as an independent measure of image--metaphor alignment. This calibration therefore
did not score any zero-shot or refined generated image: although the same underlying metaphors also appear in
the 200-metaphor evaluation set of \Cref{sec:results}, no generated image scored there was used to select \(\theta\).

\begin{table}[H]
  \centering
  \caption{Calibration of the pullback matching threshold \(\theta\) on 125 human-authored visual metaphors. \%
  Zero denotes the proportion of images for which no structural correspondence is recovered.}
  \label{tab:theta_calibration}
  \begin{tabular}{lccc}
    \toprule
    \(\boldsymbol{\theta}\) & \textbf{Mean Pullback} & \textbf{\% Zero} & \textbf{Spearman vs.\ CLIP} \\
    \midrule
    0.35 & 2.3144 & 2.4\%  & 0.0753 \\
    \textbf{0.40} & \textbf{2.2305} & \textbf{3.2\%} & 0.0578 \\
    0.45 & 2.0538 & 7.2\%  & 0.0690 \\
    0.55 & 1.4883 & 26.4\% & -0.0242 \\
    \bottomrule
  \end{tabular}
\end{table}

As \(\theta\) increases, matching becomes more restrictive: mean pullback scores decrease and zero scores
become substantially more frequent, reaching 26.4\% at \(\theta=0.55\). This is especially informative as these
results are measures on a dataset of 125 real visual metaphors, so there are relational correspondences that
should be identified in all provided images. Correlation with CLIP is weak across all settings, consistent
with CLIP measuring holistic image--text similarity rather than relational correspondence. We therefore
select \(\theta=0.40\) as a conservative operating point that retains almost all non-zero structural matches
while avoiding the more permissive matching induced by \(\theta=0.35\).

\pagebreak 

\section{Full Qualitative Comparisons}
\label{appendix:qual-comparisons}

\begin{figure}[H]
    \centering
    \includegraphics[width=1.0\linewidth]{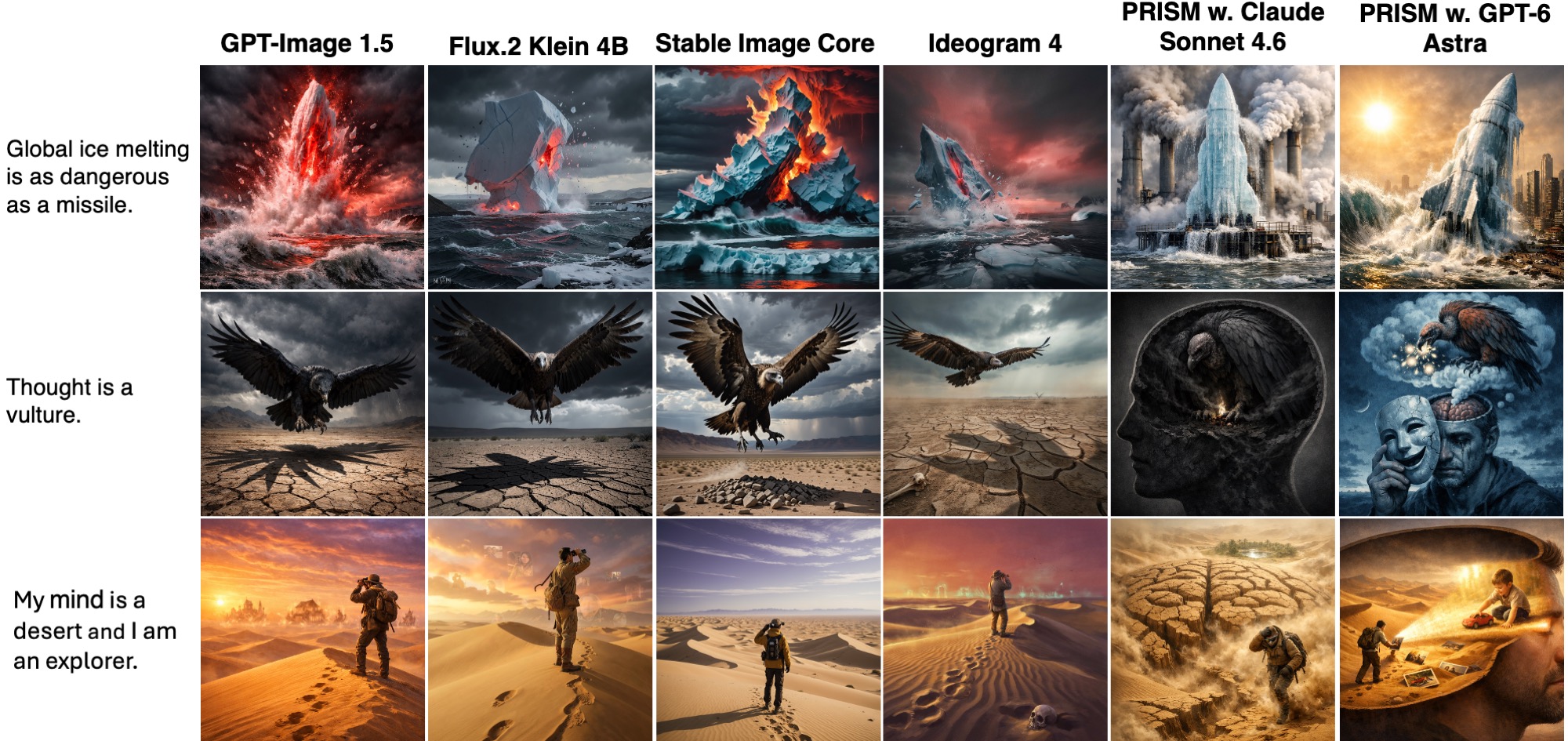}
    \caption{Comparison of four zero-shot generations pipelines with GPT-Image 1.5, Flux.2 Klein 4B, Stable Image Core and Ideogram 4 versus two versions of PRISM with Claude Sonnet 4.6 and GPT-6 Astra as the extraction and refinement models.}
    \label{fig:comparisons-full}
\end{figure}

\end{document}

%% file: math_commands.tex
\usepackage{amsmath,amsfonts,bm}

\def\eqref#1{equation~\ref{#1}}
\def\1{\bm{1}}

\DeclareMathAlphabet{\mathsfit}{\encodingdefault}{\sfdefault}{m}{sl}
\SetMathAlphabet{\mathsfit}{bold}{\encodingdefault}{\sfdefault}{bx}{n}